\documentclass[final]{article} 
\usepackage[submission]{colm2025_conference}

\usepackage{booktabs}
\usepackage{tabularx}
\usepackage{makecell}
\usepackage{multirow}
\usepackage{adjustbox}
\usepackage{siunitx}
\usepackage{xcolor}
\usepackage{rotating}

\usepackage{booktabs}
\usepackage{xcolor,colortbl}
\usepackage{adjustbox}
\usepackage{array}

\definecolor{bandA}{RGB}{214,232,247} 
\definecolor{bandB}{RGB}{221,243,234} 
\definecolor{bandC}{RGB}{255,238,214} 
\definecolor{bandD}{RGB}{237,229,252}

\definecolor{p1}{RGB}{232,244,252}
\definecolor{p2}{RGB}{214,232,247}
\definecolor{p3}{RGB}{189,217,241}
\definecolor{p4}{RGB}{160,199,234}

\renewcommand{\arraystretch}{1.12}

\usepackage{graphicx}
\usepackage{booktabs}
\usepackage{tabularx}
\usepackage{makecell}
\usepackage{adjustbox}

\newcommand{\best}[1]{\textbf{#1}}

\usepackage{microtype}
\usepackage{hyperref}
\usepackage{url}
\usepackage{graphicx}
\usepackage{amsmath}  
\usepackage{amssymb}
\usepackage{algorithm}
\usepackage{algpseudocode}

\usepackage{booktabs}
\usepackage{multirow}       
\usepackage{array}          
\usepackage{caption}        
\usepackage{adjustbox}      
\usepackage{lineno}
\usepackage{threeparttable}
\usepackage{pdflscape} 

\definecolor{darkblue}{rgb}{0, 0, 0.5}
\hypersetup{colorlinks=true, citecolor=darkblue, linkcolor=darkblue, urlcolor=darkblue}

\title{Ostrakon-VL: Towards Domain-Expert MLLM for Food-Service and Retail Stores}

\author{
  \begin{tabular}{c}
    Zhiyong Shen\textsuperscript{1}, 
    Gongpeng Zhao\textsuperscript{1}, 
    Jun Zhou, Li Yu, Guandong Kou, \\
    Jichen Li, Chuanlei Dong, Zuncheng Li, 
    Kaimao Li\textsuperscript{2}, 
    Bingkun Wei\textsuperscript{2}, 
    Shicheng Hu\textsuperscript{2}, \\
    Wei Xia\textsuperscript{3}, 
    Wenguo Duan\\
    \\
    \small Rajax Network Technology (Taobao Shangou of Alibaba) \\
  \end{tabular}
}

\begin{document}

\maketitle
\footnotetext[1]{Equal contribution.}
\footnotetext[2]{Contributed during internship.}
\footnotetext[3]{Corresponding author: \texttt{weixia.xw@alibaba-inc.com}.}

\begin{abstract}
Multimodal Large Language Models (MLLMs) have recently achieved substantial progress in general-purpose perception and reasoning. Nevertheless, their deployment in Food-Service and Retail Stores (FSRS) scenarios encounters two major obstacles: (i) real-world FSRS data, collected from heterogeneous acquisition devices, are highly noisy and lack auditable, closed-loop data curation, which impedes the construction of high-quality, controllable, and reproducible training corpora; and (ii) existing evaluation protocols do not offer a unified, fine-grained and standardized benchmark spanning single-image, multi-image, and video inputs, making it challenging to objectively gauge model robustness. To address these challenges, we first develop \textbf{Ostrakon-VL}, an FSRS-oriented MLLM based on Qwen3-VL-8B. Second, we introduce \textbf{ShopBench}, the first public benchmark for FSRS. Third, we propose \textbf{QUAD} (Quality-aware Unbiased Automated Data-curation), a multi-stage multimodal instruction data curation pipeline. Leveraging a multi-stage training strategy, Ostrakon-VL achieves an average score of \textbf{60.1} on ShopBench, establishing a new state of the art among open-source MLLMs with comparable parameter scales and diverse architectures. Notably, it surpasses the substantially larger Qwen3-VL-235B-A22B (59.4) by \textbf{+0.7}, and exceeds the same-scale Qwen3-VL-8B (55.3) by \textbf{+4.8}, demonstrating significantly improved parameter efficiency. These results indicate that Ostrakon-VL delivers more robust and reliable FSRS-centric perception and decision-making capabilities. To facilitate reproducible research, we will publicly release Ostrakon-VL and the ShopBench benchmark\footnote[4]{\url{https://github.com/Ostrakon-VL/Ostrakon-VL}}.
\end{abstract}

\section{Introduction}

For the past few years, Multimodal Large Language Models (MLLMs) have demonstrated remarkable progress in integrating visual understanding with reasoning. Models such as Qwen3-VL \citep{bai2025qwen3vltechnicalreport}, GLM-4.6V \citep{vteam2025glm45vglm41vthinkingversatilemultimodal}, GPT-5.2 \citep{openai2025gpt52}, Gemini3 \citep{google2025gemini3} and Seed 1.8 \citep{seed1_8} exemplify this trend, demonstrating strong performance in visual comprehension tasks. Nevertheless, zero-shot or few-shot applications of these MLLMs in real-world Food-Service and Retail Stores (FSRS) frequently result in suboptimal performance. For instance, despite its strong general capabilities, Qwen3-VL-235B exhibits a conspicuous robustness gap when confronted with domain-specific FSRS challenges. In tasks requiring high-fidelity reasoning, such as video authenticity verification and the identification of subtle differences between store branches, the model produces unreliable outputs. We provide a series of representative failure cases and a detailed analysis of these limitations in Appendix \ref{sec:capability_highlights}. In high-stakes operational and compliance workflows, such error rates can lead to false alarms and missed risks. We argue that this gap cannot be attributed to a single missing capability. Rather, it arises from a system-level misalignment among three factors: (i) the objectives of general-purpose MLLMs, (ii) the in-the-wild visual data distributions, and (iii) the requirements of industry for end-to-end perception-to-reasoning systems. Consequently, three primary challenges stand out.

\paragraph{Challenge 1: Capability-Level Misalignment in General-Purpose MLLMs for FSRS.}

General-purpose MLLMs exhibit a capability-level misalignment when applied to FSRS scenarios—not due to a lack of basic robustness on imperfect inputs, but because they are not equipped to interpret domain-specific visual semantics under real-world surveillance conditions. Unlike generic vision–language tasks, FSRS requires reasoning about operational cues—such as distinguishing operational signage from decorative elements—and handling characteristic degradations common in retail settings, including glare on storefront glass, low-resolution multilingual text, motion blur, and transient occlusions. Moreover, effective performance demands consistent understanding across heterogeneous subdomains (ShopFront, ShopInterior, and Kitchen) and input modalities ranging from static images to dynamic video sequences. Since these requirements are largely absent from standard MLLM pretraining and alignment data, off-the-shelf models struggle to produce reliable and semantically grounded responses in FSRS despite their general competence on broad benchmarks.

\paragraph{Challenge 2: Data-Level Noise and Heterogeneity in Real-World.}
Real-world MLLM datasets, exemplified by the FSRS domain, are inherently heterogeneous and noisy. Visual data is collected from diverse and unconstrained sources, including regulatory inspections, low-resolution surveillance cameras, and casual mobile captures, resulting in large variations in viewpoint, resolution, and environmental conditions. These factors introduce severe visual noise—such as compression artifacts, motion blur, glare, and occlusion—that significantly complicates robust perception. Beyond visual corruption, the data distribution is further destabilized by inconsistent metadata, high redundancy, and temporal drift in annotation standards. Together, these factors make stable and controllable model training and iteration particularly challenging. Consequently, a systematic and reproducible data curation pipeline is essential to transform unstructured raw observations into high-quality, diverse and balanced training signals, enabling reliable closed-loop model improvement.

\paragraph{Challenge 3: Evaluation-Level Misalignment in FSRS.}

Existing general-purpose MLLM benchmarks fail to adequately measure the core capabilities required in the FSRS domain (Challenge 1). In particular, these benchmarks neither capture model robustness to domain-specific acquisition noise nor provide sufficient granularity to assess fine-grained evidence extraction in cluttered FSRS environments. In the absence of a unified FSRS benchmark, it becomes difficult to reliably compare models or diagnose failure modes specific to FSRS. This limitation disrupts the evaluation-to-refinement loop, preventing evaluation signals from effectively guiding iterative improvements in data curation and model design.

To address these multifaceted challenges, we propose a systematic framework that integrates data curation with domain-specific expertise. Our contributions are threefold:

\begin{itemize}
     \item We introduce \textbf{Ostrakon-VL}, the first MLLM specifically tailored for the FSRS domain. By leveraging a systematic multi-stage training strategy, Ostrakon-VL (built on Qwen3-VL-8B) outperforms significantly larger general-purpose models (e.g., Qwen3-VL-235B) on \textbf{ShopBench}, while supporting robust end-to-end perception and reasoning within complex FSRS scenarios.
      \item We release \textbf{ShopBench}, a FSRS domain benchmark spanning single-image, multi-image, and video settings, enabling faithful measurement of robustness to real-world acquisition noise, fine-grained evidence extraction, multi-evidence composition, and decision consistency under explicit operational rules in real FSRS scenarios.
    \item We propose \textbf{QUAD}, a \textbf{Q}uality-driven \textbf{U}niversal \textbf{A}lignment \textbf{D}ata-curation pipeline that distills a 69.25M candidate pool into a high-signal 3.40M corpus (retaining only $\sim$1/20 of the data) while improving downstream performance.
    
\end{itemize}

\section{Related Work}

\paragraph{MLLMs.}
Recent MLLMs \citep{alayrac2022flamingo,liu2023visual,li2023blip,beyer2024paligemma,li2024llava,bai2025qwen3vltechnicalreport,vteam2025glm45vglm41vthinkingversatilemultimodal,google2025gemini3} leverage vision-LLM alignment and instruction tuning to achieve strong general-purpose perception capabilities. Subsequent works \citep{zheng2025villa,lieber2024jamba,yu2024crema,zhang2025videollama} further extend these capabilities to multi-image and video settings by optimizing data mixtures and enabling long-context reasoning. In parallel, multimodal evaluation benchmarks have rapidly evolved to assess capabilities such as Optical Character Recognition (OCR), spatial reasoning, chart/table understanding, and grounded QA, providing standardized metrics and exposing failure modes in real-world scenarios \citep{fei2025path,fu2024ocrbench}. Together, these advances establish a solid technical foundation for developing domain-specific MLLMs.

\paragraph{Domain-Specific MLLMs.}
MLLMs are driving vertical industries from ``single-modality analysis'' to a new paradigm of ``multi-source evidence fusion'' by jointly modeling text, images, tables, charts, and videos \citep{li2024multimodal}. In finance, FinTMMBench \citep{zhu2025fintmmbench} constructed an evaluation benchmark spanning time-series tables, news, and charts to characterize models' retrieval and reasoning abilities; Open-FinLLMs \citep{huang2024open} further conducted pretraining and instruction tuning on large-scale corpora integrating text, tables, time series, and charts, demonstrating stronger domain adaptation than general-purpose models. In healthcare, the benefits of domain pretraining and workflow alignment are even more pronounced: BiomedCLIP \citep{zhang2023biomedclip} aligned vision and language using large-scale biomedical image-text data and achieved strong performance on radiology-related tasks; more system-oriented efforts (e.g., BrainGPT \citep{li2025towards}) have embedded multimodal models into concrete clinical workflows, supporting 3D brain CT report generation with clinically oriented evaluations. In interdisciplinary areas such as biology and materials science, models like Cephalo \citep{buehler2024cephalo} have explored connecting biomaterial images with textual knowledge and integrating them with downstream tasks such as structure design in an end-to-end manner \citep{ghafarollahi2025automating}.

\paragraph{Multimodal data cleaning.}
High-quality and scalable data curation pipeline is key infrastructure for making MLLMs reliable and reproducible \citep{moharil2024towards,zhou2025megapairs}. \emph{Foundation-model-in-the-loop} multimodal data cleaning has become a mainstream practice in both industry and recent research. Early pipelines often rely on powerful proprietary MLLMs as high-precision ``judges'' to automatically remove noisy samples, repair weak labels, and enforce instruction/response quality at scale. On the text side, Large Language Models (LLMs) are commonly used for semantic-consistency checking, factuality and readability filtering, instruction/response quality assessment, as well as automatic correction and rewriting of weakly labeled samples \citep{zhang2023jellyfish, zhang2025coddllm}. On the vision side, MLLMs can be used for visual--textual consistency verification, visual evidence validation, and hard-example mining, improving data quality and annotation consistency while reducing human cost \citep{zhu2024vdc,braun2024defame,gu2025unime}. In addition, task-oriented data synthesis and filtering \citep{joshi2025mm} and efficient cross-modal screening for specific modality pairs such as audio--video data \citep{vosoughi2025quality} further demonstrate the broad potential of foundation models as general-purpose data governors. 

\paragraph{MLLMs for FSRS.}
In domains such as finance and healthcare, domain-specific MLLMs have gradually formed a relatively complete stack comprising \emph{domain training}, \emph{data pipelines}, and \emph{standardized evaluations}. In contrast, intelligent solutions for FSRS remain largely fragmented into task-specific systems, such as back-of-house compliance and risk monitoring, fire warning, and hygiene-score prediction \citep{wang2024kitchen,ma2020smart,wang2025predicting}. These efforts demonstrate the value of automation, yet there remains, to our knowledge, a lack of a unified MLLM, as well as reproducible data-curation pipelines and benchmarks to support general-purpose perception and reasoning across diverse FSRS scenarios.

Unlike the relatively standardized inputs in finance and healthcare, FSRS data are noisier and far more heterogeneous, making data curation and evaluation more challenging: naive MLLM-as-judge filtering is prone to spurious correlations and is difficult to audit, and it often lacks closed-loop correction based on spot checks and downstream feedback. To this end, we follow a roadmap of \emph{domain-specific model--closed-loop data pipeline--standardized evaluation}: we introduce a FSRS-focused MLLM Ostrakon-VL, build a unified data curation pipeline QUAD, and release an evaluation suite ShopBench to enable reproducible comparisons, failure-mode analysis, and iterative improvements.

\section{QUAD: Quality-aware Unbiased Automated Data-curation Pipeline}
\label{sec:quad}

\subsection{Motivation and Overview}

The instruction learning and alignment of MLLMs aim to model the conditional distribution $p(a \mid I, q)$, where $I$ represents the input image, $q$ the instruction or question, and $a$ the target answer. In this paradigm, data does not merely drive parameter updates but fundamentally sets the boundaries of the model's capabilities and its ability to generalize. However, large-scale multimodal datasets—especially those synthesized or crawled—often suffer from varying quality, ranging from factual hallucinations to heavy redundancy. 

\begin{figure}[t]
  \centering
  \includegraphics[width=\linewidth]{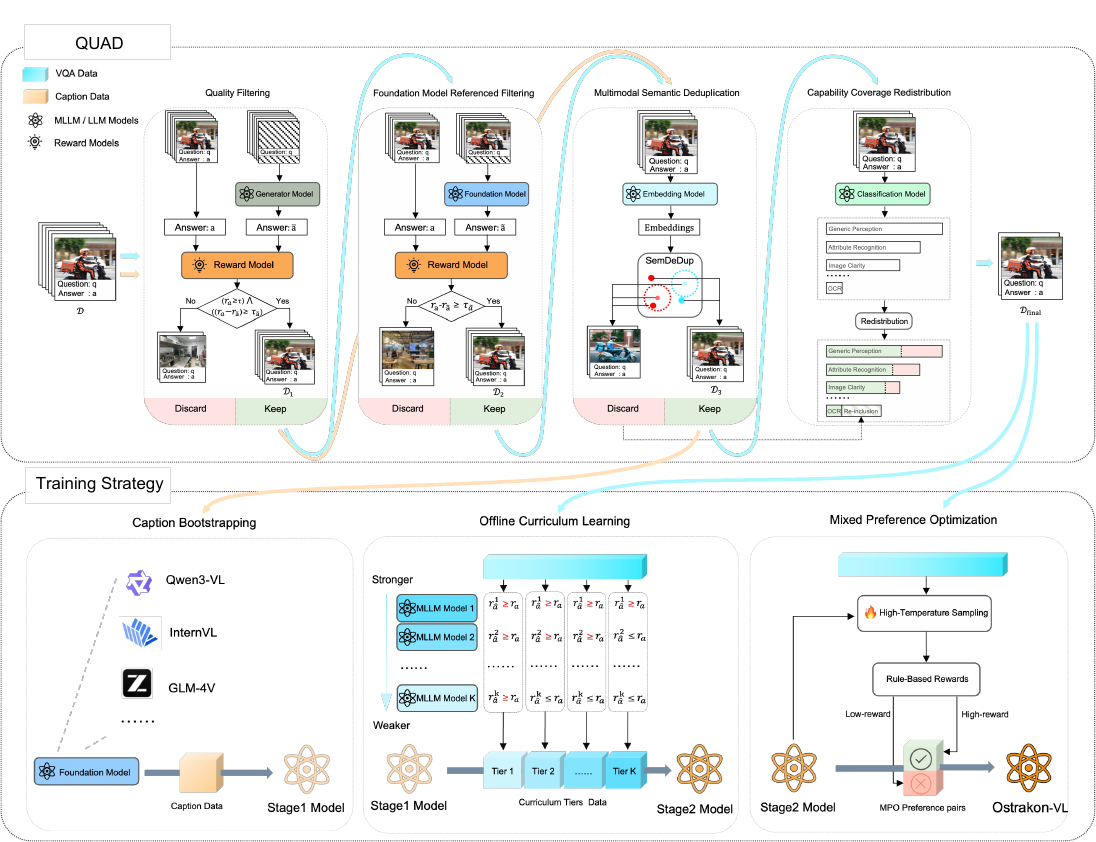}
  \caption{Overall framework of Ostrakon-VL. The framework consists of two core components: (Top) QUAD, a high-quality data curation pipeline that distills raw data into a high-signal corpus through quality filtering, foundation model referenced filtering multimodal semantic deduplication, and capability coverage redistribution; (Bottom) Training Strategy, a multi-stage process starting from domain knowledge injection via caption bootstrapping, followed by offline curriculum learning for progressive adaptation, and concluding with Mixed Preference Optimization to ensure output stability and robustness.}
  \label{fig:data-pipeline}
\end{figure}

We argue that selecting effective instruction data from a large candidate pool hinges on four cardinal factors: \textbf{Correctness}, \textbf{Learnability}, \textbf{Non-redundancy}, and \textbf{Balance}. To maximize the quality of supervision signals while maintaining task coverage and filtering low-utility samples, we propose QUAD, a multi-stage multimodal instruction data curation pipeline. As illustrated in Figure~\ref{fig:data-pipeline}, QUAD implements a systematic four-stage refinement process—comprising \emph{Quality Filtering}, \emph{Foundation Model Referenced Filtering}, \emph{Multimodal Semantic Deduplication}, and \emph{Capability Coverage Redistribution}—to progressively distill raw candidate pools into high-quality training set.

\subsection{Data Synthesis}
The primary objective of data synthesis is to transform unstructured visual resources into a vast, trainable instruction-following dataset. We employ advanced MLLMs as generators, denoted as $G_{\theta}$, taking raw visual streams and associated contextual cues—such as sample questions or descriptive captions—as inputs. By utilizing these cues as semantic anchors, the $G_{\theta}$ is prompted to perceive visual features and generate diverse instruction-response pairs that align with the provided context. This procedure facilitates the rapid construction of a large-scale preliminary corpus $\mathcal{D}$. While this synthetic approach ensures comprehensive scenario coverage, it inherently introduces "machine-originating" noise, such as vision-language hallucinations or logical inconsistencies, necessitating a rigorous and systematic curation pipeline.

\subsection{Instruction Data Filtering}
\label{sec:data_selection_cleaning}
The preliminary corpus $\mathcal{D}$ contains inevitable noise and redundancy from the automated synthesis. To resolve this, we employ QUAD to systematically distill $\mathcal{D}$ into a high-quality instruction set $\mathcal{D}_{\text{final}}$. This curation process consists of four progressive stages, which are elaborated as follows. The refinement process optimizes the dataset across four key dimensions: correctness, learnability, deduplication, and balance.
 
\paragraph{Quality Filtering.}
\label{sec:stage1}
Rule-based filters or MLLM-based consistency checks may still preserve samples that are fluent in text but weakly grounded in vision (e.g., hallucinated details or over-inference beyond visual evidence). To rigorously control data quality, we introduce a reward model \(R_{\phi}\) (Skywork-VL-Reward ~\citep{wang2025skywork_vl_reward} in our implementation) to evaluate each candidate triplet \((I,q,a)\) across dimensions of visual--textual relevance, informativeness, linguistic quality, and credibility. The reward score is defined as
\begin{equation}
r_a = R_{\phi}(I,q,a) \in \mathbb{R}.
\end{equation}
We retain samples using either a fixed threshold \(r_a \ge \tau\) or a percentile rule (e.g., Top-\(p\%\)), where \(\tau\) (or \(p\)) is selected based on validation performance and manual auditing.

To further suppress instances solvable through linguistic priors alone, we perform a vision-ablated check. Specifically, we completely omit the visual input and provide only the question $q$ to the generator $G_{\theta}$, forcing the model to generate a text-only answer based solely on textual context:

\begin{equation}
\bar{a} = G_{\theta}(q).
\end{equation}

We then compute a linguistic prior score for \((I, q,\bar{a})\) using $R_{\phi}$.
\begin{equation}
r_{\bar{a}} = R_{\phi}(I,q,\bar{a}).
\end{equation}

Intuitively, the margin $r_a - r_{\bar{a}}$ isolates the visual contribution to the quality of the answer. A small margin indicates that the original answer $a$ relies heavily on language priors rather than explicit visual evidence. Consequently, we apply a joint filtering criterion and define the refined candidate pool $\mathcal{D}_1$ as:

\begin{equation}
(I,q,a)\in\mathcal{D}_1 \iff \bigl(r_a \ge \tau\bigr)\ \wedge\ \bigl( (r_a - r_{\bar{a}}) \ge \tau_{\bar{a}}\bigr),
\end{equation}
where \(\tau_{\bar{a}}\) is a margin threshold determined by validation performance and manual auditing. This dual-track filtering ensures that $\mathcal{D}_1$ provides a factually correct and visually grounded foundation. By eliminating samples with erroneous logic or excessive linguistic bias, this stage guarantees the correctness and reliability of the supervision signals, which is essential for stable model alignment in subsequent stages.

\paragraph{Foundation Model Referenced Filtering.}
\label{sec:stage2}
High-quality samples in $\mathcal{D}_1$ do not necessarily imply high training gains: overly easy samples provide limited gradient signal, while weak labels (answers inferior to the output of foundation model) can inject harmful supervision. To estimate the marginal utility of each sample, we perform \emph{foundation model referenced filtering}: for each triplet $(I, q, a) \in \mathcal{D}_1$, we employ the foundation model to decode a reference answer $\tilde{a}$, which is then evaluated by $R_{\phi}$.
\begin{equation}
r_{\tilde{a}} = R_\phi(I,q,\tilde{a}).
\end{equation}
We define the reward gap as $\Delta r = r_a - r_{\tilde{a}}$, which serves as a proxy for the learning potential. If $\Delta r < \tau_{\tilde{a}}$, where $\tau_{\tilde{a}} \ge 0$ is a configurable filtering threshold, it indicates that the current foundation model's output $\tilde{a}$ is already better than (or sufficiently close to) $a$. Such samples are considered mastered instances that provide limited additional learning signal for the foundation model. By filtering out samples with $\Delta r < \tau_{\tilde{a}}$, we obtain a further refined set $\mathcal{D}_2$:
\begin{equation}
(I, q, a) \in \mathcal{D}_2 \iff (I, q, a) \in \mathcal{D}_1 \wedge (\Delta r \ge \tau_{\tilde{a}}).
\end{equation}
This step ensures that $\mathcal{D}_2$ contains samples that are not only factually reliable but also provide substantial learnable improvements for model alignment.

\paragraph{Multimodal Semantic Deduplication.}
\label{sec:stage3}
To increase information density and prevent the model from overfitting to redundant patterns, we perform multimodal semantic deduplication. This process aims to ensure the diversity of the dataset by pruning near-duplicate samples that offer diminishing returns.

Concretely, for each triplet $(I,q,a) \in \mathcal{D}_2$, we extract a joint vision--language embedding using a multimodal embedding model (e.g., GME-Qwen2VL-2B ~\citep{zhang2024gme}). Following the \textsc{SemDeDup} framework~\citep{abbas2023semdedup}, we first apply k-means clustering in the embedding space to partition the data into semantic clusters, and then perform intra-cluster deduplication via a k-center style selection procedure to remove near-duplicates while preserving semantic coverage.

This hierarchical selection process preserves the semantic coverage of the data while effectively pruning redundant information, yielding the non-redundant set $\mathcal{D}_3$. By maximizing the uniqueness and diversity of each sample, this stage ensures that the model is exposed to a wide range of unique multimodal scenarios.

\paragraph{Capability Coverage Redistribution.}
\label{sec:stage4}
The cumulative effect of multi-stage filtering may inadvertently shift the data distribution, potentially leading to an imbalance across different capability axes. While generic visual tasks often remain abundant, specialized skills—such as OCR, relational reasoning, or complex spatial analysis—risk being marginalized if the selection process remains unconstrained. To preempt such distributional skews and ensure a robust balance across the model's capability surface, we implement a taxonomy-driven redistribution strategy to produce the final dataset $\mathcal{D}_{\text{final}}$.

By synthesizing established vision benchmarks and the practical functional requirements of MLLMs, we derive both a systematic capability taxonomy $\mathcal{C}$ and a target prior distribution $\pi(c)$ that reflects the desired balance across scenarios. To map the instances in $\mathcal{D}_3$ into this taxonomy, we employ a task-specific capability classifier $M_\mathcal{C}$, which can be instantiated by any sufficiently powerful pre-trained LLM. In our implementation, $M_\mathcal{C}$ is fine-tuned from Qwen3-8B using a curated seed set of approximately 7,000 high-fidelity, crowdsourced instances. For each question $\in \mathcal{D}_3$, the classifier predicts its functional category as:
\begin{equation}
c = M_\mathcal{C}(q).
\end{equation}

Based on these categorical assignments, we perform stratified re-sampling to align the data distribution with the pre-defined prior $\pi(c)$, yielding the final curated dataset $\mathcal{D}_{\text{final}}$. This step ensures a balanced and representative distribution across all functional domains, effectively mitigating the bias toward high-frequency tasks. A detailed comparison of capability coverage before and after redistribution is provided in Appendix~\ref{sec:vqa_curation}.

\section{Training Strategy}
We propose a systematic, domain-agnostic multi-stage training strategy (Figure~\ref{fig:data-pipeline}) to enhance multimodal learning. While broadly applicable, it proves especially beneficial for complex domains like FSRS that demand fine-grained perception and reasoning. First, we perform Caption Bootstrapping (CB)—a domain-knowledge injection stage in which the foundation model is trained on FSRS images paired with dense, evidence-rich captions. By applying rigorous \emph{Quality Filtering} and \emph{Multimodal Semantic Deduplication}, we ensure the resulting caption supervision is factually grounded, free of hallucinations, and semantically diverse, thereby establishing a robust foundation for downstream "intent–evidence–answer" alignment. Second, we employ Offline Curriculum Learning (OCL) to mitigate the optimization instability typically caused by significant domain shifts. By stratifying and scheduling instruction data according to difficulty, this stage facilitates a smooth transition from general-purpose perception to domain-specific expertise. Finally, the model undergoes Mixed Preference Optimization (MPO)~\citep{wang2024enhancing}, where we construct challenging preference pairs between correct responses and plausible but incorrect alternatives to sharpen fine-grained discrimination and improve response reliability. Through this progressive pipeline, Ostrakon-VL evolves from fundamental domain grounding to sophisticated task alignment and robust decision-making behavior.

\subsection{Caption Bootstrapping}
To inject domain knowledge in FSRS scenarios, we employ a CB stage before downstream instruction tuning. In this phase, the foundation model is trained on FSRS images paired with dense captions that explicitly describe FSRS domain visual evidence, such as signage text, menu boards, equipment, dining layout, and kitchen cues. This design is motivated by recent studies showing that high-quality caption supervision substantially improve multimodal grounding and provide richer evidence tokens for subsequent instruction following and reasoning tasks \citep{chen2024sharegpt4v,10655294,chen2024far,alves2024artificial}. In particular, caption supervision provides information-rich labels that capture multiple entities, attributes, and relations in a single sample — an advantage that is particularly important under the domain shifts commonly encountered in FSRS.

Before training, we refine the raw caption pool by applying the QUAD (see Sec.~\ref{sec:quad}), focusing on two high-impact stages: (i) \emph{Quality Filtering}, where a reward model scores visual–textual consistency and grounding to remove hallucinated or weakly evidenced captions; and (ii) \emph{Multimodal Semantic Deduplication}, which removes near-duplicate image--caption pairs in a joint vision--language embedding space, preventing repetitive templates from dominating optimization and improving effective coverage. For the captioning corpus, we bypass the \emph{Foundation Model Referenced Filtering} and \emph{Capability Coverage Redistribution} stages. This decision stems from the real-world FSRS caption data, where noise is dominated by corruption, factual errors, and templated phrasing rather than descriptions being too simplistic. Furthermore, as captioning data lacks a structured question-answer (QA) format, it cannot be categorized into specific capability domains for redistributing. By focusing on quality filtering and deduplication, we improve caption reliability without substantially narrowing the data distribution, establishing a robust foundation for the "intent-evidence-answer" alignment in subsequent VQA training.

\subsection{Offline Curriculum Learning}
Curriculum Learning (CL) posits that organizing training data from easy to difficult enables models to learn more efficiently, mirroring the human learning process \citep{bengio2009curriculum}. 
For domain-specific reasoning in FSRS, directly optimizing models on the full dataset—containing samples of varying difficulty and supervision quality—can lead to suboptimal convergence or unstable training. To alleviate this problem, we introduce an offline difficulty stratification strategy that progressively guides learning from basic patterns to complex reasoning tasks.

To enable reproducible OCL, we estimate sample difficulty via model reasoning with a set of reference MLLMs. For each triplet $(I,q,a) \in \mathcal{D}_{\text{final}}$, we employ $K$ diverse, off-the-shelf MLLMs $\{M_k\}_{k=1}^K$ to generate candidate answers $\hat{a}^{k} \sim M_k(I,q)$. We then use a reward model $R_\phi$ to evaluate the quality of both the target answer $a$ and the generated responses:

\begin{equation}
r_a = R_\phi(I,q,a), \qquad r^k_{\hat{a}} = R_\phi(I,q,\hat{a}_{k}).
\end{equation}
We define the per-model reward gap as
\begin{equation}
\Delta r^k_{\hat{a}} = r^k_{\hat{a}} - r_a,
\end{equation}
and then count the number of reference models whose generated answers strictly outperform the ground-truth answer by a margin threshold $\tau_{cl}$:
\begin{equation}
s = \sum_{k=1}^{K} \mathbb{I}\!\left[\Delta r^k_{\hat{a}} > \tau_{cl}\right].
\end{equation}
Based on the resulting score $s$, we stratify the $\mathcal{D}_{\text{final}}$ into $n$ curriculum tiers, where $n\in\mathbb{N}$ and $1\le n\le K+1$, ordered from easier to harder:
\begin{equation}
\text{Tier}(I,q,a)= f(s), \qquad f:\{0,1,\ldots,K\}\rightarrow \{1,\ldots,n\}.
\end{equation}

Intuitively, if many reference models can readily surpass the provided supervision, the instance is likely easy to learn. In contrast, for which most reference models produce low-quality responses provide more challenging yet informative supervision signals. This voting-based criterion is simple, auditable, and fully reproducible given $(\{M_k\}, R_\phi, \tau_{cl}, f)$.

\subsection{Mixed Preference Optimization}
Inspired by the proven success of MPO in the Intern-VL series ~\citep{chen2024internvl, wang2025internvl35advancingopensourcemultimodal, zhu2025internvl3} and subsequent models such as Skywork R1V2 ~\citep{wang2025skywork} and Keye-VL ~\citep{team2025kwai, yang2025kwai}, we adopt it as a core component of the Ostrakon-VL training strategy. MPO is specifically selected for its superior alignment effectiveness and high computational efficiency. In particular, compared with Group Relative Policy Optimization (GRPO)~\citep{shao2024deepseekmath}, MPO is significantly more computationally efficient, as it avoids the substantial memory and time overhead associated with online group sampling and large-scale reward computation.

The MPO objective consists of three complementary components, enabling the model to learn from multiple perspectives. The overall loss is defined as a weighted sum:
\begin{equation}
    \mathcal{L}_{MPO} = w_1 \mathcal{L}_{preference} + w_2 \mathcal{L}_{quality} + w_3 \mathcal{L}_{generation}
\end{equation}

where each weight $w_*$ controls the contribution of its corresponding term. The preference loss, derived from DPO ~\citep{rafailov2023direct} trains the model to distinguish which of two candidate responses is preferable. The quality loss, inspired by BCO ~\citep{jung2025binary}, encourages the model to assess the intrinsic quality of an individual response. The generation loss acts as a regularization term to preserve fluency and stability, while preventing excessive deviation from the pretrained distribution.

This integrated design enables Ostrakon-VL to achieve high-quality alignment while maintaining training efficiency. Compared to GRPO-style methods, MPO substantially reduces computational overhead by eliminating the need to generate and evaluate multiple candidate responses per input during training. Moreover, by jointly optimizing preference, quality, and generation objectives, MPO mitigates a common failure mode of standard DPO where the model may unintentionally reduce the probability of producing any response — including correct ones. This leads to significantly more stable and reliable training dynamics.

The effectiveness of MPO is highly dependent on the quality of preference pair construction. We therefore adopt high-temperature sampling to produce multiple responses for each input, revealing inconsistent behaviors in which the model alternates between correct and incorrect predictions across trials, instead of exhibiting uniformly correct or incorrect outputs. These instances are especially valuable, as sporadic correctness often reflects superficial pattern matching rather than a true understanding of the underlying reasoning.

To enhance robustness, we construct preference pairs by pairing each input’s highest-quality correct response with a corresponding plausible but incorrect alternative generated by the model. Specifically, we employ rule-based rewards to differentiate correct responses from plausible but incorrect ones. These plausible but incorrect responses are nearly accurate but contain subtle errors—such as miscounted objects or minor factual inaccuracies. By explicitly contrasting its high-quality erroneous outputs with correct ones, the model learns to attend to fine-grained visual and semantic details that were previously overlooked. This strategy substantially improves output stability and accuracy, leading to more reliable performance in FSRS scenarios.

\section{ShopBench: A Benchmark Focused on the FSRS Domain}
\begin{figure}[t]
  \centering
  \includegraphics[width=\linewidth]{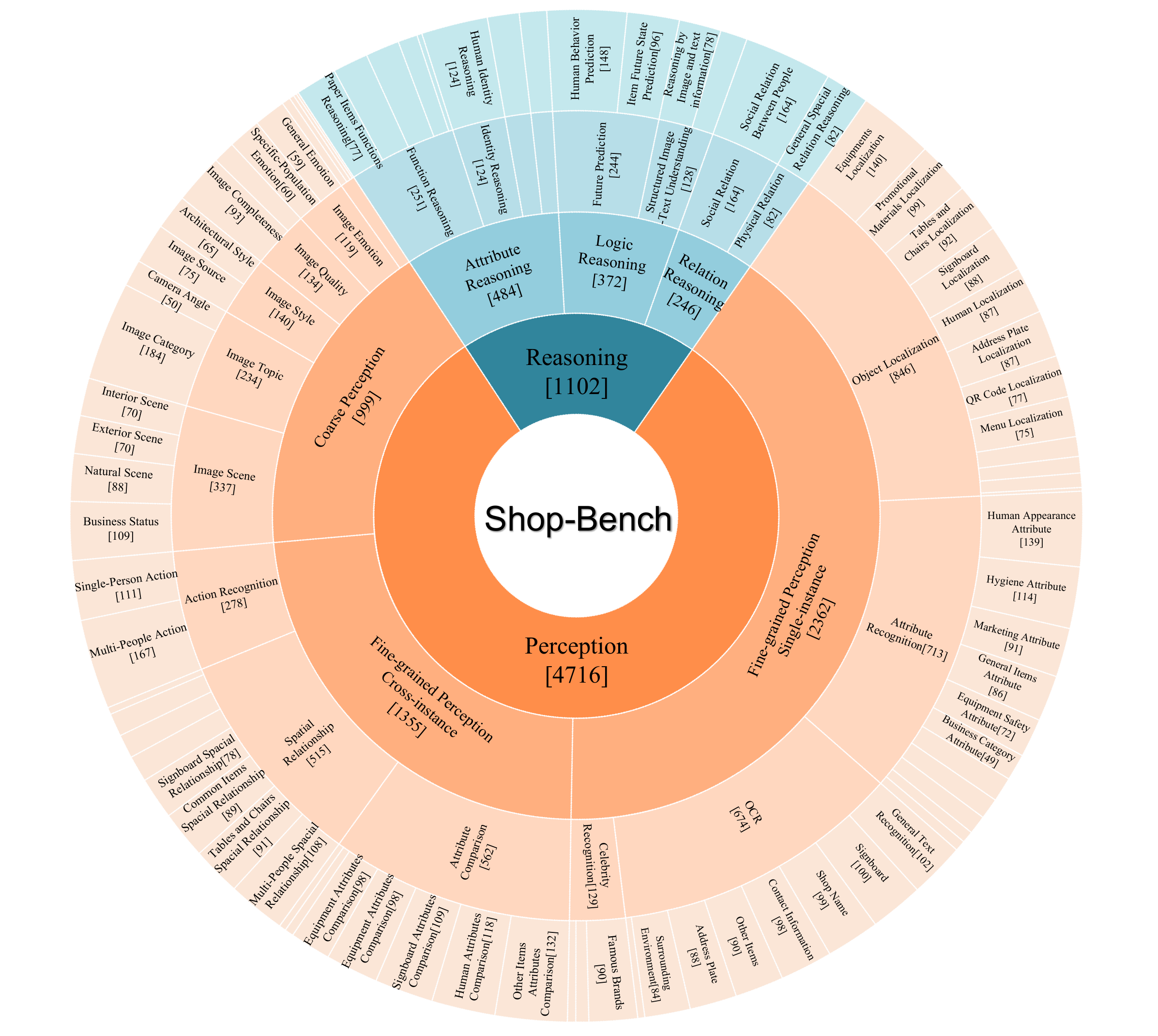}
  \caption{ShopBench taxonomy (L1--L4). We unify diverse tasks labels into a canonical L4 label space for consistent evaluation. For visual clarity, L4 categories with limited sample counts are omitted from the figure.}
  \label{fig:shopbench_taxonomy}
\end{figure}

\begin{figure}
    \centering
    \includegraphics[width=1.0\linewidth]{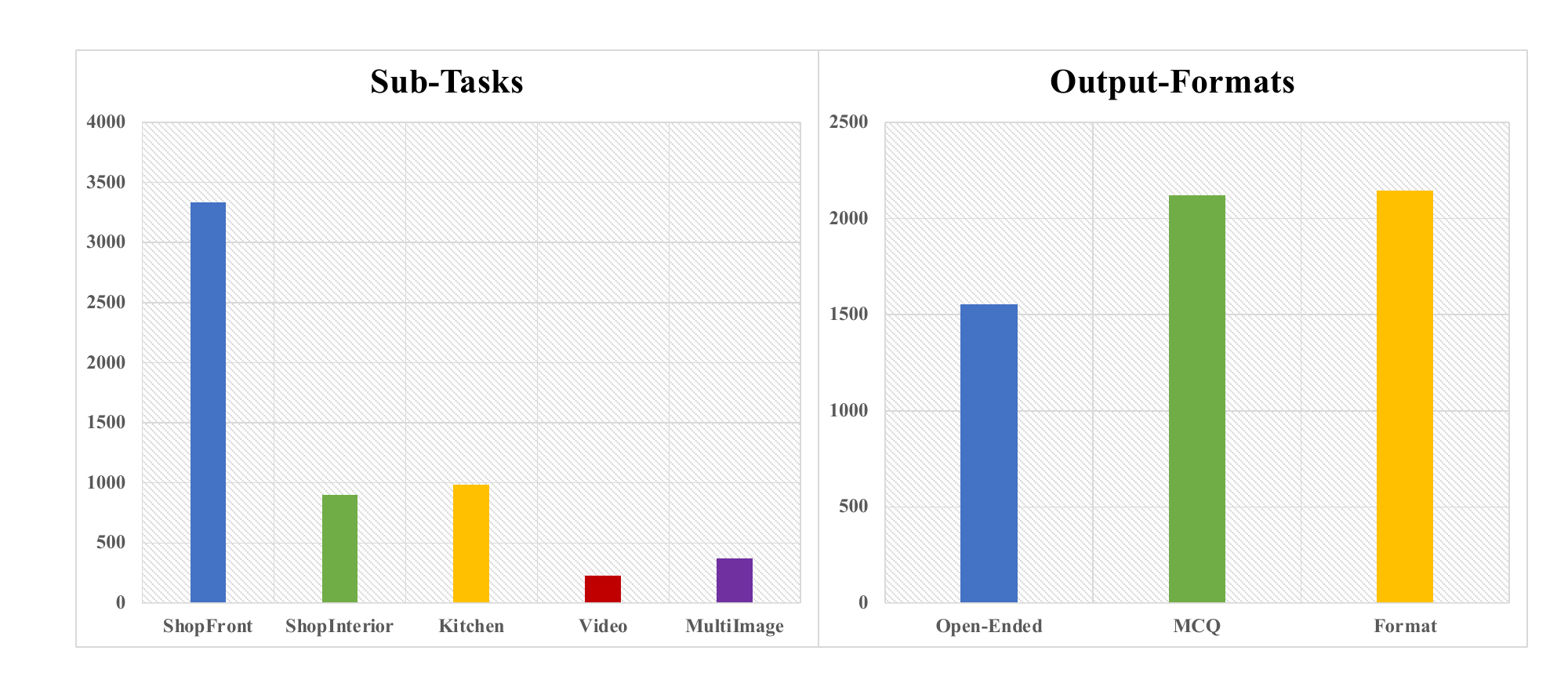}
    \caption{Counts and Proportions of Samples across Input Subtasks and Output Formats}
    \label{fig:counts_and_prop}
\end{figure}

To overcome the limitations of existing benchmarks in evaluating FSRS-specific capabilities, we present ShopBench, a specialized evaluation benchmark tailored to the FSRS domain. ShopBench is designed to assess the fine-grained perception and reasoning capabilities that are essential for retail compliance and operational decision-making. ShopBench covers three scenario categories—ShopFront, ShopInterior, and Kitchen—and three visual input formats: single-image, multi-image, and video. For clearer and more fine-grained analysis of model capability across heterogeneous settings, we structure ShopBench into five subtasks: \textbf{ShopFront}, \textbf{ShopInterior}, \textbf{Kitchen}, \textbf{MultiImg} (multi-image mixed scenarios), and \textbf{Video} (video mixed scenarios). This partition facilitates interpretable reporting by disentangling scenario-specific competence from input-format effects, and enables more diagnostic comparisons across diverse deployment conditions. Additionally, to accommodate diverse application forms and deployment flexibility, as illustrated in Figure~\ref{fig:counts_and_prop}, we design three output formats: \textbf{Open-Ended} for free-form question answering, \textbf{Format} for question answering with predefined output schemas, and \textbf{MCQ} for multiple-choice responses.

ShopBench differs from existing benchmarks in three key aspects: (1) While prior benchmarks may support multiple input formats, ShopBench is the first to organize single-image (ShopFront, ShopInterior, Kitchen), multi-image (MultiImg), and video (Video) scenarios into a unified, fine-grained hierarchical taxonomy specifically designed for diagnostic evaluation in the FSRS domain; (2) ShopBench targets the FSRS domain, resulting in an image-domain distribution that is substantially different from that of prior public benchmarks; and (3) ShopBench evaluation protocol emphasizes the necessity of visual evidence in multimodal assessment, thereby reducing measurement bias arising from knowledge leakage or over-reliance on language priors. Below we will delve into more details of the construction of ShopBench.

\subsection{ShopBench Taxonomy}
\label{app:task_taxonomy}

To provide a comprehensive evaluation of MLLMs in the FSRS domain, ShopBench organizes all tasks under a rigorous four-level hierarchical taxonomy (L1–L4). As depicted in Figure~\ref{fig:shopbench_taxonomy}, this taxonomy systematically maps each question from broad capability families down to 79 granular task definitions at the leaf level. At the foundational level (L1), the benchmark is divided into two capability families: perception and reasoning. The perception branch distinguishes between coarse perception and fine-grained perception, with the latter further subdivided into single-instance and cross-instance settings to evaluate the model’s spatial understanding and comparative reasoning. Simultaneously, the reasoning branch is structured into attribute, logical, and relational reasoning, to evaluate the model’s ability to interpret complex FSRS scenarios. At the leaf level (L4), we define a standardized taxonomy of fine-grained evaluation categories that is agnostic to input format, harmonizing tasks across visual inputs—such as business status from a single image and action recognition from a video stream—under a unified framework. By mapping all tasks to this unified L4 framework, ShopBench enables consistent and interpretable evaluation, facilitating direct comparison of model capabilities across image-based (single/multi-image) and video-based settings.

\subsection{Domain Distinctiveness and Scene Complexity}

\begin{figure}[t]
    \centering
    \includegraphics[width=0.8\linewidth]{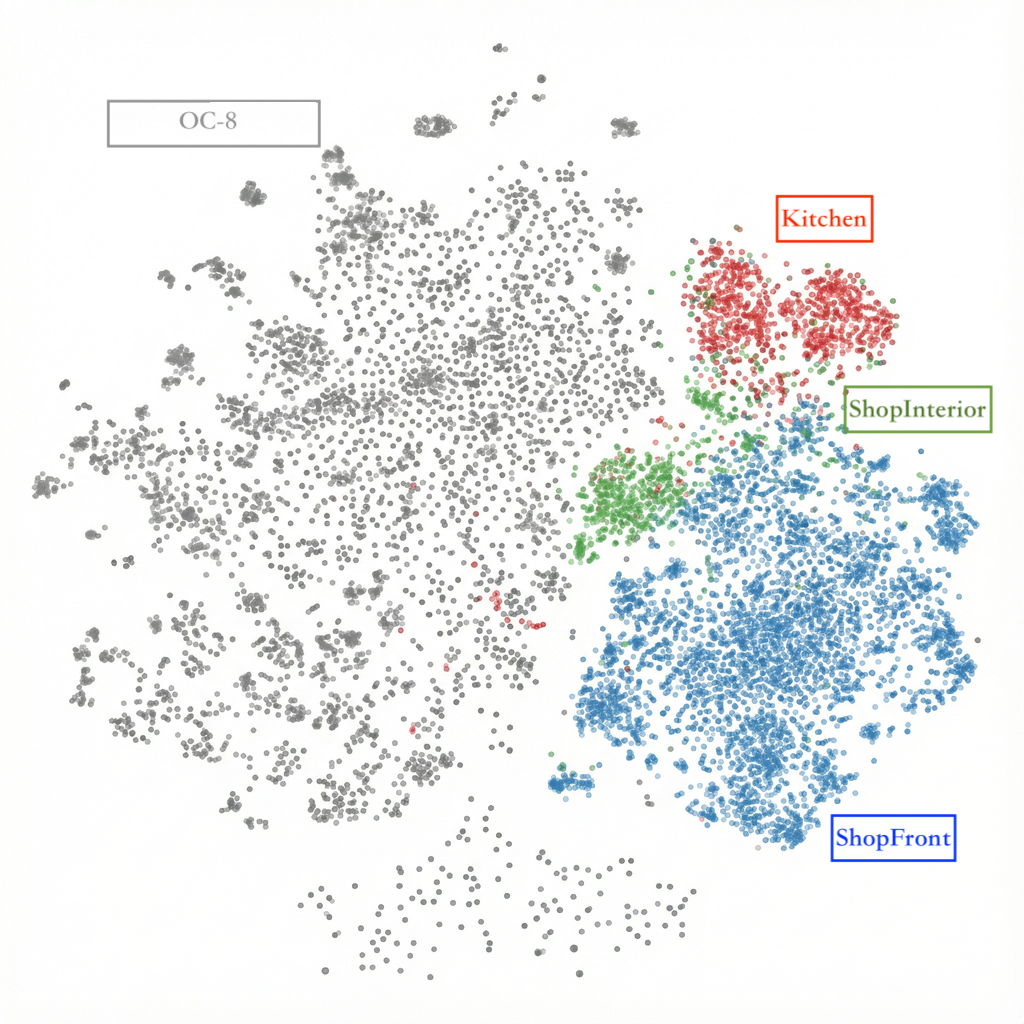}
    \caption{Distribution of different Benchmarks. OC-8(Opencompass prevailing used eight benchmarks) and ShopBench(ShopFront, ShopInterior and Kitchen)}
    \label{fig:embeddings}
\end{figure}

To visualize the coverage of existing benchmarks and illustrate how ShopBench expands the evaluation landscape, we compare its visual domain distribution and semantic scope against those of current mainstream multimodal benchmarks.

We use the GME-Qwen2VL-2B model to extract high-dimensional image embeddings from eight widely adopted benchmarks—MMBench-EN \citep{liu2024mmbench}, MMStar \citep{chen2024we}, MMVet \citep{yu2023mm}, HallusionBench \citep{guan2024hallusionbench}, AI2D \citep{hiippala2021ai2d}, OCRBench \citep{liu2024ocrbench}, MathVista \citep{lu2023mathvista}, and MMMU \citep{yue2024mmmu}—alongside ShopBench. These embeddings are projected into a two-dimensional space using t-SNE \citep{maaten2008visualizing} for comparative visualization. As illustrated in Figure~\ref{fig:embeddings}, ShopBench forms a largely distinct cluster in the embedding space, with limited overlap with existing benchmarks. This separation suggests that ShopBench captures a unique visual and semantic distribution reflective of FSRS environments, thereby meaningfully extending the current evaluation landscape for MLLMs.

Beyond domain distribution, we further quantify scene complexity by measuring the average number of object instances per image (InsPerImg) as a proxy for visual information density. Using the Recognize Anything Model \citep{2024Recognize}, we compute InsPerImg across all benchmarks. As shown in Table~\ref{tab:ins_per_img}, ShopBench exhibits the highest average instance density per image, indicating rich intra-image object interactions and more demanding scene understanding requirements.

This remarkably high instance density reflects the inherent complexity of FSRS environments, characterized by crowded retail shelves, extensive SKU variability, and intricate human-object interactions. Such visual clutter poses a substantial challenge for MLLMs, demanding not only precise fine-grained perception but also the ability to reason about numerous co-occurring objects within a single scene.

\begin{table}[t]
\centering
\setlength{\tabcolsep}{6pt}
\caption{Number of Instances per Image(InsPerImg)}
\label{tab:ins_per_img}
\begin{tabular}{l c}
\toprule
\textbf{Benchmark} & \textbf{InsPerImg} \\
\midrule
MMBench-EN      & 10.6 \\
MMStar          & 9.3  \\
MMvet           & 9.6  \\
HallusionBench  & 5.8  \\
AI2D            & 7.9  \\
OCRBench        & 6.2  \\
MathVista        & 7.1  \\
MMMU            & 6.1  \\
ShopBench       & 13.0  \\
\bottomrule
\end{tabular}
\end{table}

\subsection{Beyond Multimodal Gain: Diagnostic Evaluation Metrics}
In evaluating MLLMs, apparent performance gains can be misleading. As highlighted by MMStar \citep{chen2024we}, these gains often stem from:
(i) Visual Irrelevance: questions solvable through linguistic priors or internal knowledge;
(ii) Unintentional Leakage: answers inferred from spurious cues in prompts or metadata.
While multimodal Gain (MG) and multimodal Leakage (ML) are commonly used to measure the contribution of visual input, 
we argue that MG may not fully capture genuine visual grounding and inherently favors evaluation sets with relatively simple semantic structures.

\paragraph{The Limitations of MG.} Following MMStar, MG is defined as:
\begin{equation}
\mathrm{MG} = \mathrm{Acc}_{V}-\mathrm{Acc}_{-V} = \frac{|S_{V}|}{N} - \frac{|S_{-V}|}{N}.
\label{eq:mg}
\end{equation}

and ML is defined as:
\begin{equation}
\mathrm{ML} = max(0, \mathrm{Acc}_{-V}-\mathrm{Acc}_{T}) = max(0, \frac{|S_{-V}|}{N} - \frac{|S_{T}|}{N}).
\label{eq:mg}
\end{equation}

where $N$ is the total number of instances in the benchmark, ${S}_{V}$ and ${S}_{-V}$ denote the sets of instances correctly answered with and without visual input, respectively. ${S}_{T}$ denote the sets of instances correctly answered with MLLM’s LLM base (without any multi-modal training).

From a benchmark design perspective, the primary limitation of MG is its lack of diagnostic granularity regarding task difficulty and scene complexity.

First, MG can create a misleading impression of multimodal necessity in benchmarks dominated by strong language priors. In such cases, a high MG may reflect the benchmark’s semantic simplicity—where models succeed by relying solely on linguistic cues—rather than genuine visual grounding.

Second, in visually complex domains, MG suffers from a suppression effect: even when tasks are inherently vision-dependent, extreme visual clutter can cause models to perform worse with visual input than without it. This results in reduced or negative MG values that mask the benchmark’s true capacity to evaluate sophisticated multimodal understanding.

Consequently, MG cannot reliably distinguish whether its value stems from genuine reliance on visual input or from performance degradation caused by visual clutter. This ambiguity renders MG an inadequate metric for assessing a benchmark’s diagnostic depth.

\paragraph{Two diagnostic metrics.} 
To address the limitations of MG, we decompose it into two complementary metrics: Visual Necessity Rate (VNR), which quantifies the proportion of instances where visual input is indispensable for a correct response, and Vision-Induced Failure (VIF), which measures the extent to which visual information interferes with predictions that would otherwise be correct based on textual cues alone. Let $\mathcal{D}_{\text{eval}}$ be an evaluation set with $N$ instances and $\mathcal{F}_{V}:=\mathcal{D}_{\text{eval}} \setminus S_{V}$ represent the set of failures when vision is provided, the definitions of VNR and VIF are given by the following formulas:

\begin{equation}
\mathrm{VNR} = \frac{|S_{V}\setminus S_{-V}|}{|S_{V}|}.
\label{eq:vnr}
\end{equation}

\begin{equation}
\mathrm{VIF} = \frac{|\mathcal{F}_{V} \cap S_{-V}|}{|\mathcal{F}_{V}|}.
\label{eq:vif}
\end{equation}

Together, VNR and VIF offer a more granular and interpretable characterization of how models rely on visual input, disentangling genuine visual necessity from detrimental visual interference. As normalized metrics bounded in $[0,1]$, they allow fair and intuitive comparison between benchmarks—unlike MG, which conflates these opposing effects. For more details, see Appendix~\ref{sec:details_of_metrics}.

\begin{table*}[t]
\centering
\small
\setlength{\tabcolsep}{3pt}
\renewcommand{\arraystretch}{1.12}
\caption{Comparison on open-source and domain benchmarks. Each benchmark includes VNR, VID, and ML metrics. 
Per-metric top-2 and top-5 scores are indicated in \textbf{bold} and \underline{underlined}. }
\label{tab:open_domain_vnr_vif_ml}
\sisetup{table-alignment=center}
\begin{adjustbox}{width=1.0\textwidth, center}
\begin{tabular}{%
l
@{\hspace{5pt}}
S[table-format=1.2] S[table-format=1.2] S[table-format=1.2] S[table-format=1.2]
@{\hspace{5pt}}
S[table-format=1.2] S[table-format=1.2] S[table-format=1.2] S[table-format=1.2]
@{\hspace{5pt}}
S[table-format=1.2] S[table-format=1.2] S[table-format=1.2] S[table-format=1.2]
@{\hspace{5pt}}
S[table-format=1.2] S[table-format=1.2] S[table-format=1.2] S[table-format=1.2]
@{\hspace{5pt}}|
S[table-format=1.2] S[table-format=1.2] S[table-format=1.2] S[table-format=1.2]
}
\toprule

\multirow{2}{*}{Benchmark} &
\multicolumn{4}{c}{\makecell[c]{Qwen3-VL-235B}} &
\multicolumn{4}{c}{\makecell[c]{Qwen3-VL-8B}} &
\multicolumn{4}{c}{\makecell[c]{InternVL3.5-8B}} &
\multicolumn{4}{c}{\makecell[c]{MiniCPM-V2.6}} & 
\multicolumn{4}{|c}{\makecell[c]{AVG}} \\
\cmidrule(lr){2-5}\cmidrule(lr){6-9}\cmidrule(lr){10-13}\cmidrule(lr){14-17}\cmidrule(lr){18-21}
& {\textsc{VNR$\uparrow$}} & {\textsc{VIF$\downarrow$}} & {\textsc{MG$\uparrow$}} & {\textsc{ML$\downarrow$}}
& {\textsc{VNR$\uparrow$}} & {\textsc{VIF$\downarrow$}} & {\textsc{MG$\uparrow$}} & {\textsc{ML$\downarrow$}}
& {\textsc{VNR$\uparrow$}} & {\textsc{VIF$\downarrow$}} & {\textsc{MG$\uparrow$}} & {\textsc{ML$\downarrow$}}
& {\textsc{VNR$\uparrow$}} & {\textsc{VIF$\downarrow$}} & {\textsc{MG$\uparrow$}} & {\textsc{ML$\downarrow$}} 
& {\textsc{VNR$\uparrow$}} & {\textsc{VIF$\downarrow$}} & {\textsc{MG$\uparrow$}} & {\textsc{ML$\downarrow$}} 
\\
\midrule

\textbf{MMBench-EN} &
\underline{0.81} & \underline{0.08} & \best{0.72} & 0.66 &
\underline{0.85} & 0.11 & \best{0.62} & \underline{0.03} &
\underline{0.85} & \underline{0.03} & \best{0.68} & \underline{0.04} &
\underline{0.79} & \underline{0.05} & \best{0.59} & 0.53 &
\underline{0.83} & \underline{0.07} & \best{0.66} & 0.32 \\

\textbf{MMStar} &
\underline{0.70} & \underline{0.11} & \underline{0.49} & \underline{0.28} &
\underline{0.86} & \underline{0.06} & \underline{0.51} & \underline{0.02} &
\underline{0.64} & \underline{0.16} & \underline{0.29} & \underline{0.06} &
\underline{0.73} & \underline{0.13} & \underline{0.25} & \best{0.00} &
\underline{0.73} & \underline{0.12} & \underline{0.41} & \underline{0.09} \\

\textbf{MMVet} &
\best{0.92} & \best{0.03} & \underline{0.53} & 0.50 &
\best{0.95} & \best{0.02} & \underline{0.39} & 0.05 &
\best{0.93} & \best{0.01} & \underline{0.31} & 0.10 &
\best{0.86} & \best{0.04} & 0.24 & 0.08 &
\best{0.92} & \best{0.03} & \underline{0.37} & 0.18 \\

\textbf{HallusionBench} &
0.37 & 0.16 & 0.20 & \best{0.00} &
\underline{0.80} & 0.11 & 0.22 & \underline{0.02} &
0.51 & 0.20 & 0.06 & 0.10 &
0.25 & 0.21 & 0.22 & 0.07 &
0.48 & 0.17 & 0.17 & \best{0.05} \\

\textbf{AI2D} &
0.26 & 0.21 & 0.19 & \underline{0.19} &
0.48 & \underline{0.08} & \underline{0.37} & 0.05 &
0.24 & 0.19 & 0.15 & \best{0.03} &
0.32 & 0.26 & 0.20 & \best{0.00} &
0.33 & 0.19 & 0.23 & \underline{0.07} \\

\textbf{OCRBench} &
\best{0.99} & \best{0.00} & \best{0.69} & \underline{22} &
\best{0.97} & \best{0.00} & \best{0.65} & \best{0.00} &
\best{0.97} & \best{0.00} & \best{0.58} & 0.04 &
\best{0.97} & \best{0.00} & \best{0.65} & \best{0.00} &
\best{0.98} & \best{0.00} & \best{0.64} & \underline{0.07} \\

\textbf{MathVista} &
0.52 & 0.17 & \underline{0.36} & 0.77 &
0.62 & 0.12 & 0.38 & 0.12 &
0.52 & 0.16 & 0.21 & \underline{0.05} &
0.56 & 0.16 & \underline{0.25} & \underline{0.02} &
0.56 & 0.15 & \underline{0.30} & 0.24 \\

\textbf{MMMU} &
0.35 & 0.23 & 0.19 & 0.71 &
0.60 & 0.16 & 0.24 & 0.08 &
0.53 & 0.21 & 0.20 & 0.12 &
0.51 & 0.25 & 0.17 & 0.15 &
0.50 & 0.21 & 0.20 & 0.27 \\

\textbf{ShopBench} &
\underline{0.63} & \underline{0.13} & 0.27 & \best{0.07} &
0.56 & \underline{0.11} & 0.23 & \best{0.00} &
\underline{0.55} & \underline{0.12} & \underline{0.25} & \best{0.02} &
\underline{0.61} & \underline{0.13} & \underline{0.28} & \best{0.00} &
\underline{0.58} & \underline{0.12} & 0.25 & \best{0.02} \\

\bottomrule
\end{tabular}
\end{adjustbox}
\end{table*}

Detailed metrics—including VNR, VIF, MG, and ML—on general and domain-specific benchmarks are reported in Table~\ref{tab:open_domain_vnr_vif_ml}. While MG varies noticeably across models due to differences in scale or architecture, VNR and VIF remain remarkably consistent, suggesting they are primarily determined by the benchmark itself rather than model-specific factors.

On OCRBench, the model achieves an almost perfect VNR (1.0) and VIF (0.0). This is expected, as OCR is essentially a vision-only task: VNR and VIF—designed to detect VQA-specific issues such as language priors or visual interference—naturally approach their theoretical baselines (VNR = 1.0, VIF = 0.0) when the task relies solely on visual input and involves little external knowledge or linguistic priors. In contrast, MG—which reflects multimodal gain rather than leakage—ranks OCRBench only second, behind the general-purpose MMBench, contrary to expectations. This discrepancy further underscores the diagnostic value of VNR and VIF over MG in characterizing true visual dependence. Additionally, MG does not provide an intuitive indication of benchmark quality. 

MMVet ranks second across both average VNR (0.92) and VIF (0.03), trailing only OCRBench, indicating that its dataset design emphasizes visually grounded evaluation and effectively minimizes spurious cues that could mislead model reasoning. Similar trends are observed on MMBench, MMStar, and ShopBench, which also exhibit high average VNR and low average VIF, reflecting careful construction to promote genuine visual grounding. Notably, ShopBench shows the lowest average ML (0.02), as expected, further confirming its resistance to knowledge leakage. Together, these benchmarks provide a comprehensive and fair arena for evaluating current MLLMs.

\section{Experiments}
In this section, we first evaluate Ostrakon-VL on ShopBench against current state-of-the-art (SoTA) MLLMs. 
We then assess its general-purpose capabilities on widely adopted public benchmarks, comparing it with peer MLLMs of comparable scale. Finally, we present a comprehensive ablation study of our QUAD data pipeline and training strategy,  isolating the contribution of each component in both data construction and model training. All training and evaluation parameters are detailed in Section~\ref{sec:train_eval_details}.

\subsection{Comparison with the SOTA Models on ShopBench}

We benchmark Ostrakon-VL against representative MLLMs across scales—from 4B to 241B—including open models from the Qwen-VL, InternVL, GLM, LLaVA-OneVision~\citep{li2024llavaOneVision}, MiniCPM ~\citep{yao2024minicpm}, and Ovis ~\citep{lu2025ovis25technicalreport} families, as well as closed-source systems (Gemini, GPT-5, Seed). All comparisons use public checkpoints or official APIs under unified evaluation settings.

\paragraph{Benchmarking mainstream MLLMs on ShopBench.}
As evidenced in Table \ref{tab:shop_scene_benchmarks}, 
Ostrakon-VL establishes strong performance in the FSRS domain, achieving an average score of 60.1 on ShopBench—the highest among all 8B-scale open models and 4.8 points higher than its Qwen3-VL-8B baseline. It ranks second overall among all ranked open-source models, behind only the 241B InternVL3.5-241B (61.1), yet outperforms dense models nearly 9× larger such as Qwen2.5-VL-72B (57.3) and surpasses the much larger Qwen3-VL-235B MoE (59.4), demonstrating exceptional parameter efficiency. A granular subtask analysis reveals that Ostrakon-VL excels in critical FSRS scenarios—achieving the highest score among all ranked open-source models in ShopFront (65.0) and the second-highest in Kitchen (59.1). These results indicate that general-purpose scaling alone is insufficient for precise understanding in the diverse and complex settings of FSRS. Instead, the integration of domain-specific instructions (FSRS-Inst) and our training strategies empowers Ostrakon-VL—a compact 8B model—to achieve top-ranked performance on key subtasks, offering a highly efficient and effective solution for real-world FSRS visual QA.

\begin{table}[t]
\centering
\caption{Performance Comparison on ShopBench, Close-source models are marked with an “*”. They are reported for reference only and are excluded from the final performance ranking. For models with the same parameter scale as Ostrakon, the corresponding rows are highlighted in blue. The highest and second-best scores are shown in \textbf{bold} and \underline{underlined} respectively}
\label{tab:shop_scene_benchmarks}
\begin{adjustbox}{width=\textwidth,center}
\begin{tabular}{l c c | c c c c c | c}
\toprule
\textbf{Metric}  & \textbf{Param} & \textbf{Architecture} & \textbf{ShopFront} & \textbf{ShopInterior} & \textbf{Kitchen} & \textbf{MultiImg} & \textbf{Video} & \textbf{AVG} \\
\midrule
MiniCPM-V-4     & 4B  & Dense & 49.2 & 42.3 & 44.0 & 36.9 & 43.6 & 46.2 \\
Qwen3-VL-3B   & 3B & Dense & 56.9 & 44.6 & 53.8 & 47.9 & 49.8 & 53.6 \\
InternVL3.5-4B   & 4B & Dense & 50.0 & 44.3 & 48.7 & 39.3 & 44.5 & 47.9 \\
\midrule
\rowcolor{bandA}
LLava-OneVision   & 7B & Dense & 37.5 & 39.2 & 41.6 & 27.1 & 35.2 & 37.6 \\
\rowcolor{bandA}
InternVL3.5-8B    & 8B & Dense & 52.0 & 46.3 & 50.1 & 41.3 & 47.5 & 49.9 \\
\rowcolor{bandA}
Qwen3-VL-8B     & 8B & Dense & 59.3 & 46.9 & 54.1 & 46.2 & 50.7 & 55.3 \\
\rowcolor{bandA}
GLM-4.6V-FlashX    & 9B & Dense & 61.9 & 47.6 & 57.0 & 56.5 & 47.1 & 57.9 \\
\rowcolor{bandA}
Ovis2.5-9B     & 9B & Dense & 61.8 & 47.6 & 56.4 & \underline{57.3} & 49.8 & 57.9 \\
\midrule
Qwen2.5-VL-72B  & 72B & Dense  & 60.8 & 49.5 & 57.3 & 49.3 & 50.7 & 57.3 \\
InternVL3-78B   & 78B & Dense  & 58.3 & 51.0 & 54.8 & 49.2 & 49.8 & 55.7 \\
GLM-4.6V   & 106B & MoE & \underline{64.0} & \underline{51.1} & 57.3 & 57.0 & \underline{54.1} & 60.0 \\
Qwen3-VL-235B   & 235B & MoE & 63.2 & 50.3 & 58.6 & 53.1 & 53.3 & 59.4 \\
InternVL3.5-241B & 241B & MoE & 63.5 & \best{52.7} & \best{62.9} & \best{59.7} & \best{54.2} & \best{61.1} \\
\midrule
Gemini2.5-Pro* & -- & MoE & 52.2 & 56.5 & 65.9 & 61.5 & 56.7 & 55.9 \\
Gemini3-Pro* & -- & MoE & 64.7 & 70.6 & 53.4 & 65.8 & 62.3 & 63.6 \\
Seed 1.8* & -- & MoE & 73.8 & 57.9 & 63.5 & 70.1 & 63.9 & 69.0 \\
GPT-5*  & -- & MoE & 57.2  & 56.2 & 65.3 & 69.0 & 48.3 & 58.8 \\
\midrule
\rowcolor{bandA}
Ostrakon-VL  & 8B & Dense & \best{65.0} & 49.1 & \underline{59.1} & 49.6 & 53.3 & \underline{60.1} \\
\bottomrule
\end{tabular}
\end{adjustbox}
\end{table}

\subsection{Performance on Public Multimodal Benchmarks}

\begin{table}[t]
\centering
\caption{Performance comparison of multimodal models on representative benchmarks. The highest and second-best scores are shown in \textbf{bold} and \underline{underlined} respectively.}
\label{tab:multimodal_performance}
\begin{adjustbox}{width=\textwidth,center}
\begin{tabular}{lcccccc}
\toprule
\textbf{Dataset} &
\textbf{Qwen2.5-VL-7B} &
\textbf{Qwen3-VL-8B} &
\textbf{InternVL3.5-8B} &
\textbf{GLM-4.6V-FlashX} &
\textbf{Ovis2.5-9B} &
\textbf{Ostrakon-VL-8B} \\
\midrule

\multicolumn{7}{c}{\textbf{Comprehensive multimodal understanding \& hallucination}} \\
MMBench-EN$_{DEV}$      & 81.8 & 83.9 & 81.6 & \underline{84.2} & \best{84.3} & 82.0 \\
MMStar          & 60.8 & 67.7 & 62.8 & \best{70.6} & \underline{68.6} & 61.1 \\
MMVet           & 59.2 & \underline{61.2} & 50.6 & 61.0 & \best{62.2} & 36.4 \\
HallusionBench  & 47.8 & \underline{57.1} & 51.5 & 56.6 & \best{60.5} & 54.0 \\
\midrule

\multicolumn{7}{c}{\textbf{OCR-related understanding}} \\
AI2D$_{TEST}$          & 80.6 & 84.5 & 79.7 & \best{85.0} & \underline{84.6} & 83.0 \\
OCRBench        & \underline{85.4} & 85.3 & 78.4 & 85.3 & \best{85.9} & 61.7 \\
DocVQA          & 93.3 & \underline{94.4} & 90.6 & 92.0 & \best{95.1} & 85.7 \\
\midrule

\multicolumn{7}{c}{\textbf{Math \& knowledge}} \\
MathVista$_{MINI}$       & 68.8 & \underline{77.2} & 75.3 & \best{82.5} & 67.7 & 75.4 \\
MMMU$_{VAL}$           & 51.5 & \underline{57.1} & 54.4 & 54.2 & \best{60.6} & 54.8 \\
\midrule

\multicolumn{7}{c}{\textbf{Chinese-language ability}} \\
MMBench-CN$_{DEV}$       & \best{88.2} & 83.5 & 81.0 & 84.2 & \underline{85.0} & 80.3 \\
Chinese-OCRBench & \underline{88.6} & \best{90.6} & 80.4 & 85.4 & 84.6 & 88.5 \\
CMMMU            & 33.0 & 33.1 & \best{40.0} & 36.3 & \best{40.0} & \underline{33.2} \\
CMATH            & 74.8 & 66.2 & 62.0 & \best{84.8} & \underline{83.4} & 63.5 \\
\midrule
AVG       & 70.3 & 72.4 & 68.3 & \underline{74.0} & \best{74.0} & 66.7 \\
\bottomrule
\end{tabular}
\end{adjustbox}
\end{table}

We evaluate Ostrakon-VL on a range of public vision-language benchmarks against state-of-the-art open-source models of comparable scale (7B–9B), including Qwen2.5-VL-7B ~\citep{bai2025qwen25vltechnicalreport}, Qwen3-VL-8B, InternVL3.5-8B, GLM-4.6V-FlashX and Ovis2.5-9B. 

Our evaluation protocol spans four key dimensions: (i) General Comprehension and Robustness, assessed on MMBench-EN, MMStar, and MMVet for scene understanding, alongside HallusionBench to detect hallucinatory tendencies; (ii) OCR-related Capabilities, evaluated on AI2D, OCRBench, and DocVQA ~\citep{mathew2021docvqa}; (iii) Complex Reasoning, measured by MathVista and MMMU; and (iv) Chinese Multimodal Understanding, tested on MMBench-CN ~\citep{liu2024mmbench}, Chinese-OCRBench ~\citep{liu2024ocrbench}, CMMMU ~\citep{zhang2024cmmmu}, CMATH ~\citep{wei2023cmath}. This multi-faceted diagnostic ensures a holistic assessment of whether domain-specific optimization compromises foundational multimodal competence.

As shown in Table~\ref{tab:multimodal_performance}, Ostrakon-VL exhibits a clear trade-off between domain specialization and general multimodal competence. Its average score (66.7) is notably lower than that of the base model Qwen3-VL-8B (72.4), reflecting the effect of focused fine-tuning on FSRS-specific data. However, this degradation is not uniform: Ostrakon-VL retains strong performance on tasks aligned with FSRS scenarios—such as Chinese-OCRBench (88.5, close to the best 90.6)—and maintains reasonable capability in mathematical reasoning (MathVista: 75.4) and hallucination control (HallusionBench: 54.0). Crucially, this modest reduction in general benchmarks is offset by a substantial gain (+4.8 points) on ShopBench, our target domain. The results suggest that Ostrakon-VL successfully prioritizes task-relevant skills without catastrophic forgetting of foundational abilities, striking a practical balance for real-world FSRS deployment.

\subsection{Ablation Study}
In this section, we conduct extensive ablation studies to systematically examine the individual contributions of the key components in Ostrakon-VL. Our analysis focuses on two core aspects of the system-level design: the QUAD pipeline for data curation and the multi-stage training strategy. All experiments are evaluated on ShopBench to ensure a rigorous and fair assessment of the effectiveness of each design choice.

\subsubsection{Ablation on QUAD.}
The results reported in Table~\ref{tab:indomain_pipeline_config} demonstrate the efficiency of our four-stage QUAD pipeline in distilling high-quality instruction data. Notably, the pipeline achieves an overall compression ratio of $20.4\times$, reducing the initial dataset from 69.25M samples to a refined set of 3.40M, while improving average performance from 56.7 to 59.2—a net gain of 2.5 points. This improvement underscores a fundamental principle in domain-specific fine-tuning: a highly distilled, task-aligned dataset is substantially more effective than a massive but noisy corpus. These findings highlight that prioritizing data quality and semantic relevance enables superior model performance with less than 5\% of the original data volume.

It is important to emphasize that evaluations at each stage are conducted independently per subtask category to ensure robust performance across diverse scenarios. In the \emph{Quality Filtering} stage, we achieve a compression ratio of $4.2\times$ (16.40M samples), raising the average score to 58.2. This is followed by \emph{Foundation Model Referenced Filtering}, which further compresses the dataset to 8.10M samples ($8.6\times$) and improves performance to 58.9. The third stage, \emph{Multimodal Semantic Deduplication}, achieves the highest compression ratio of $23.4\times$  (2.96M samples). While this leads to a slight drop in average score to 58.4, the decline is likely due to the removal of semantically redundant samples—instances that, despite their repetitiveness, may have incidentally reinforced model robustness through broad (though inefficient) coverage of edge cases. Nevertheless, this pruning is essential to enhance instructional efficiency and eliminate redundancy. Finally, the \emph{Capability Coverage Redistribution} stage adjusts the task distribution to 3.40M samples ($20.4\times$ compression), achieving the peak average of 59.2. These findings confirm that redistributing capability coverage is a vital final step in unlocking the full potential of a distilled, high-quality instruction set.

\begin{table}[htbp]
\centering
\small
\setlength{\tabcolsep}{2.5pt} 
\renewcommand{\arraystretch}{1.2}
\caption{Stage-wise configuration of our four-stage QUAD pipeline on the in-domain dataset. The "C.R." column denotes the Compression Ratio relative to the Raw Data volume (69.25\textbf{M}illon), calculated as $Original / Current$. '--' indicates values inherited from the previous valid stage.}
\label{tab:indomain_pipeline_config}
\begin{adjustbox}{width=\textwidth,center}
\begin{tabular}{lccccccc|c}
\toprule
\textbf{Stage} & \textbf{Total} & \textbf{C.R.} & \textbf{ShopFront} & \textbf{ShopInterior} & \textbf{Kitchen} & \textbf{MultiImg} & \textbf{Video}  & \textbf{AVG} \\
\midrule
Raw Data & 69.25M & -- & 61.9 & 45.4 & 56.1 & 47.2 & 45.3 & 56.7 \\

\textit{+ Quality Filtering} & 16.40M & 4.2$\times$ & 63.2 & 46.9 & 56.5 & 51.2 & 49.7 & 58.2 \\
\textit{+ Foundation Model Referenced Filtering} & 8.10M & 8.6$\times$ & 63.6 & 48.0 & 58.2 & 48.8 & 53.3 & 58.9 \\
\textit{+ Multimodal Semantic Deduplication} & 2.96M & 23.4$\times$ & 62.6 & 47.7 & 58.2 & 50.4 & -- & 58.4 \\
\textit{+ Capability Coverage Redistribution} & 3.40M & 20.4$\times$ & 63.6 & 48.2 & 59.2 & -- & -- & 59.2 \\
\bottomrule
\end{tabular}
\end{adjustbox}
\end{table}

\subsubsection{Ablation on training strategies.}

We ablate our multi-stage training strategy: (i) \textbf{CB}, which warms up the model with dense FSRS captions for domain grounding; (ii) \textbf{OCL}, which schedules instruction tuning from easy to hard based on offline difficulty stratification; and (iii) \textbf{MPO}, which further aligns the model using preference pairs between correct responses and plausible-but-incorrect alternatives.

The ablation results in Table \ref{tab:ablation_study_on_training_strategy} evaluate the contribution of our training strategies (CB, OCL, and MPO) on top of the FSRS-Inst-enhanced baseline (ID 2). Starting from the foundation model (ID 1, AVG = 55.3), the addition of FSRS-Inst (a domain-specific VQA instruction set constructed using the QUAD pipeline, with its internal data randomly mixed during training) improves performance to 56.8 (ID 2), confirming the value of domain-specific instructions. Specifically, CB continues training the base model (ID 1) on the dense FSRS captions. In contrast, OCL uses the same FSRS-Inst data and initialization (ID 1), but organizes the data according to sample difficulty, enabling curriculum-guided learning from simpler to more complex instances. All data components underlying FSRS-Inst, as well as those used in CB, OCL, and MPO, are detailed in Appendix~\ref{sec:train_data_curation}. All subsequent variants (IDs 3–9) build upon ID 2, enabling a clean assessment of each strategy in isolation (IDs 3–5), in pairs (IDs 6–8), and all together (ID 9).

Evaluated individually on the FSRS-Inst baseline (ID 2), OCL (ID 4, AVG = 57.9) yields the largest gain (+1.1), with notable improvements on Kitchen and Video, suggesting that difficulty-aware curriculum learning aids complex visual reasoning; CB (ID 3, AVG = 57.4) also boosts performance, especially on ShopFront, indicating that bootstrapped textual refinements enhance visual-language alignment for ShopFront understanding; in contrast, MPO (ID 5, AVG = 57.0) provides only marginal overall improvement, though it slightly increases ShopFront accuracy by 0.6 points. Crucially, pairwise combinations are highly complementary: CB+OCL (ID 8, 59.3) greatly outperforms either alone, with CB’s semantics enhancing OCL’s curriculum; CB+MPO (ID 6, 58.5) and OCL+MPO (ID 7, 58.5) similarly exceed their single-strategy baselines, confirming MPO’s additive benefit. The full Ostrakon-VL model (ID 9) achieves the best average score (60.1), outperforming the base model by 4.8 points and the FSRS-Inst baseline by 3.3 points. It leads on four of five tasks, demonstrating that the joint integration of our training strategies is essential for strong performance in FSRS visual QA.

\begin{table}[htbp]
\centering
\caption{Ablation study of our training strategies. \textbf{Base} refers to the Qwen3-VL-8B, and \textbf{FSRS-Inst} denotes our domain-specific VQA instruction set, while \textbf{CB}, \textbf{OCL}, and \textbf{MPO} represent our proposed training strategies.}
\label{tab:ablation_study_on_training_strategy}
\begin{adjustbox}{width=\textwidth,center}
\begin{tabular}{c l ccc ccccc | c}
\toprule
\textbf{ID} & \textbf{Variant} & \textbf{CB} & \textbf{OCL} & \textbf{MPO} & \textbf{ShopFront} & \textbf{ShopInterior} & \textbf{Kitchen} & \textbf{MultiImg} & \textbf{Video} & \textbf{AVG} \\
\midrule
1 & Base          &  &  &  & 59.3 & 46.9 & 54.1 & 46.2 & 50.7 & 55.3 \\
2 & + FSRS-Inst   &  &  &  & 60.9 & 47.9 & 55.3 & 48.3 & 52.4 & 56.8 \\
\midrule
3 & \multirow{3}{*}{Individual} 
              & \checkmark &  &  & 61.7 & 47.7 & 56.0 & 49.1 & 52.4 & 57.4 \\
4 &           &  & \checkmark &  & 62.2 & 47.8 & 56.9 & 48.8 & 53.8 & 57.9 \\
5 &           &  &  & \checkmark & 61.5 & 48.2 & 55.4 & 46.9 & 50.2 & 57.0 \\
\midrule
6 & \multirow{3}{*}{Combined} 
              & \checkmark &  & \checkmark & 63.5 & 47.8 & 57.0 & 48.0 & 51.6 & 58.5  \\
7 &           &  & \checkmark & \checkmark & 63.2 & 47.6 & 57.1 & 49.6 & 53.3 & 58.5 \\
8 &           & \checkmark & \checkmark &  & 64.0 & 48.9 & 57.5 & 49.1 & \best{55.6} & 59.3 \\
\midrule
9 & Ostrakon-VL  & \checkmark & \checkmark & \checkmark & \best{65.0} & \best{49.1} & \best{59.1} & \best{49.6} & 53.3 & \best{60.1} \\
\bottomrule
\end{tabular}
\end{adjustbox}
\end{table}

\section{Conclusion}

This paper investigates a practically important yet underexplored setting: deploying MLLMs in FSRS environments.
FSRS applications pose unique robustness challenges due to in-the-wild visual corruptions, heterogeneous input formats (single-image, multi-image, and video), and domain-specific decision rules that require consistent, auditable reasoning.
To address these challenges, we present an end-to-end and reproducible framework consisting of a domain-expert MLLM (\textbf{Ostrakon-VL}), a standardized evaluation benchmark (\textbf{ShopBench}), and a closed-loop data curation pipeline (\textbf{QUAD}).
Our results demonstrate the effectiveness of the QUAD pipeline and our multi-stage training strategy in building FSRS expertise: built on the same Qwen3-VL-8B backbone, Ostrakon-VL achieves 60.1 on ShopBench, outperforming the base model by +4.8 and surpassing the substantially larger Qwen3-VL-235B (59.4) by +0.7.
Moreover, QUAD reduces the instruction corpus by over 95\% while improving performance by 2.5 points, indicating that high signal-to-noise supervision and calibrated capability coverage matter more than sheer data volume.
We will release Ostrakon-VL and ShopBench to facilitate reproducible comparisons and iterative improvements; future work will extend stronger temporal and rule-centric evaluations, and incorporate stricter auditing protocols and uncertainty modeling to further enhance reliability.

\section{Acknowledgments}
We would like to sincerely thank 
Changming Li,
Chunyuan Xu,
Ercong Cheng,
Feiyan Tan,
Guanzheng Jiang,
Hao Wu,
Haodong Ru,
Huinan Zhang,
Jianbo Chen,
Jianfei Xiao,
Jianfeng Li,
Jiaping Deng,
Jiaxiong Mei,
Jiayu He,
Jing Guo,
Kaidi Chen,
Kaipeng Huang,
Liang Tang,
Long Zhang,
Maomei Liu,
Menghan Ye,
Min Zhou,
Nana Jia,
Qiao Wu,
Rendao Cao,
Tao Cui,
Tiansen Guo,
Wanghua Luo,
Weiyang Zhang,
Xiaofeng Chen,
Xiaorui Zhu,
Xin Yi,
Xingfeng Gu,
Xinyu Yang,
Yabo Gao,
Yangyang Gao,
Yanjie Zhou,
Yansong Zhou,
Yi Duan,
Yinying Zhang,
Yiying Zhang,
Yongyao Zhang,
Yue Zhang,
Yufang Lu,
Yuneng Ye,
Yuqi Huang,
Zenglong Liu,
Zhanpeng Fu,
Zhouping Yang,
Zijie Li,
Zuozhan Ding,
for their insightful discussions and unwavering support. Their valuable contributions have been instrumental in advancing the development, evaluation, defect analysis, production deployment of Ostrakon-VL, as well as its future research directions.

\bibliography{main.bib}
\bibliographystyle{IEEEtran}

\clearpage
\appendix
\section*{Appendix}
This appendix provides additional details omitted from the main paper due to space limits,
including (i) data curation and corpus statistics (\S A), (ii) ShopBench composition (\S B),
and (iii) training and evaluation settings (\S C), as well as qualitative capability examples (\S D).

\section{Dataset Corpus Curation}
\subsection{Training Data Curation}
\label{sec:train_data_curation}
The performance and robustness of our model rely on a carefully curated multimodal dataset. In this section, we detail our data preparation pipeline, which progresses from foundational visual-linguistic alignment to scenario-specific refinement. We first describe the caption and VQA instruction data used for basic multimodal grounding. To integrate these heterogeneous data types and support stable training, we present the tiered data organization within the OCL framework. Finally, we detail the data distribution for MPO, which is designed to align the model with domain-specific requirements and enhance decision-making reliability.

\subsubsection{Caption Data}
As shown in Table \ref{tab:caption_stats}, the caption corpus is exclusively constructed for three single-image FSRS scene types: ShopFront, ShopInterior, and Kitchen. Through the QUAD pipeline, 25.71M raw samples are distilled into a high-quality set of 4.23M. \emph{Quality Filtering} accounts for the majority of this reduction, effectively removing large-scale noise—particularly in the ShopFront data. Subsequently, \emph{Multimodal Semantic Deduplication} eliminates redundant samples while preserving semantic diversity, yielding a compact yet representative corpus for domain-specific training.

\begin{table}[htbp]
\centering
\small
\setlength{\tabcolsep}{10pt} 
\caption{Progression of data volume for the caption corpus throughout QUAD across single-image scenarios.} 
\begin{tabular}{lccc|c}
\toprule
\textbf{Stage} & ShopFront & ShopInterior & Kitchen & Total \\
\midrule
Raw Data & 18.51M & 6.00M & 1.20M & 25.71M \\
+ Quality Filtering & 2.31M & 3.50M & 0.92M & 6.73M \\
+ Multimodal Semantic Deduplication & 1.81M & 1.50M & 0.92M & 4.23M \\
\bottomrule
\end{tabular}
\label{tab:caption_stats}
\end{table}

\subsubsection{VQA Instruction Data}
\label{sec:vqa_curation}

Table \ref{tab:vqa_stats} details the systematic distillation of the VQA instruction corpus via the QUAD pipeline. Starting from 69.25M raw samples, the process produces a refined set of 3.4M high-quality instructions. The \emph{Quality Filtering} stage drives the largest reduction, particularly in the Kitchen data, by removing low-quality and corrupted instances. Subsequently, \emph{Foundation Model Referenced Filtering} significantly reduces the volume by retaining only samples with high instructional value. \emph{Multimodal Semantic Deduplication} then eliminates redundancy while preserving semantic diversity, yielding 2.96M samples. Finally, \emph{Capability Coverage Redistribution} adjusts the data distribution based on preliminary model performance across key reasoning capabilities to mitigate capability-specific biases and enhance targeted coverage of underperforming skills.

\begin{table}[htbp]
\centering
\small
\setlength{\tabcolsep}{4pt} 
\caption{Progression of data volume for the VQA instruction corpus throughout QUAD.}
\begin{adjustbox}{width=\textwidth,center}
\begin{tabular}{lccccc|c}
\toprule
\textbf{Stage} & ShopFront & ShopInterior & Kitchen & MultiImg & Video & Total \\
\midrule
Raw Data                  & 12.55M & 6.50M & 40.00M & 10.00M & 0.20M & 69.25M \\
+ Quality Filtering         & 4.36M  & 3.50M & 2.70M  & 5.76M  & 0.08M & 16.40M\\
+ Foundation Model Referenced Filtering   & 0.97M  & 1.40M & 1.60M  & 4.05M  & 0.08M & 8.10M  \\
+ Multimodal Semantic Deduplication & 0.66M  & 0.80M & 0.88M  & 0.54M  & -- & 2.96M\\
+ Capability Coverage Redistribution & 0.90M  & 1.00M & 0.88M  & --  & -- & 3.40M\\
\bottomrule
\end{tabular}
\end{adjustbox}
\label{tab:vqa_stats}
\end{table}

\begin{figure}[t]
    \centering
    \includegraphics[width=\linewidth]{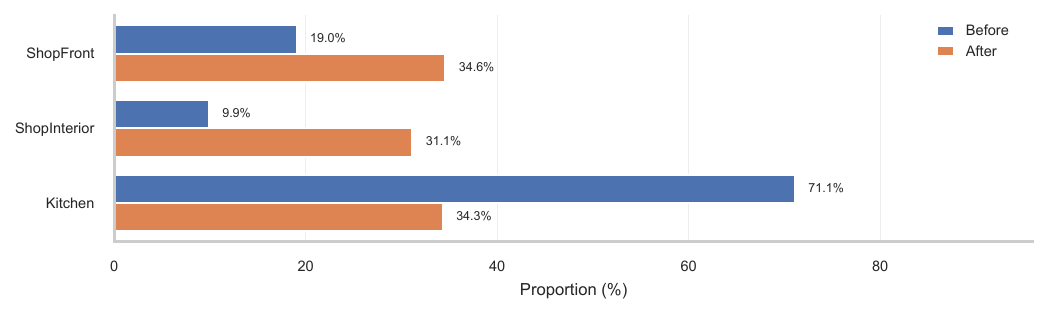}
    \caption{Data distribution across three subsets before and after the redistribution stage.}
    \label{fig:sft_subset_distribution_ws_quad}
\end{figure}

The impact of \emph{Capability Coverage Redistribution} in our QUAD pipeline is quantified in Figure~\ref{fig:sft_subset_distribution_ws_quad}. The original instruction corpus (blue) is heavily skewed toward the Kitchen subset (71.1\%), with ShopFront (19.0\%) and ShopInterior (9.9\%) significantly underrepresented. After redistribution (orange), the composition becomes substantially more balanced: Kitchen is reduced to 34.3\%, while ShopFront and ShopInterior increase to 34.6\% and 31.1\%, respectively, mitigating bias inherited from the original data distribution.

\begin{figure}[t]
    \centering
    \includegraphics[width=\linewidth]{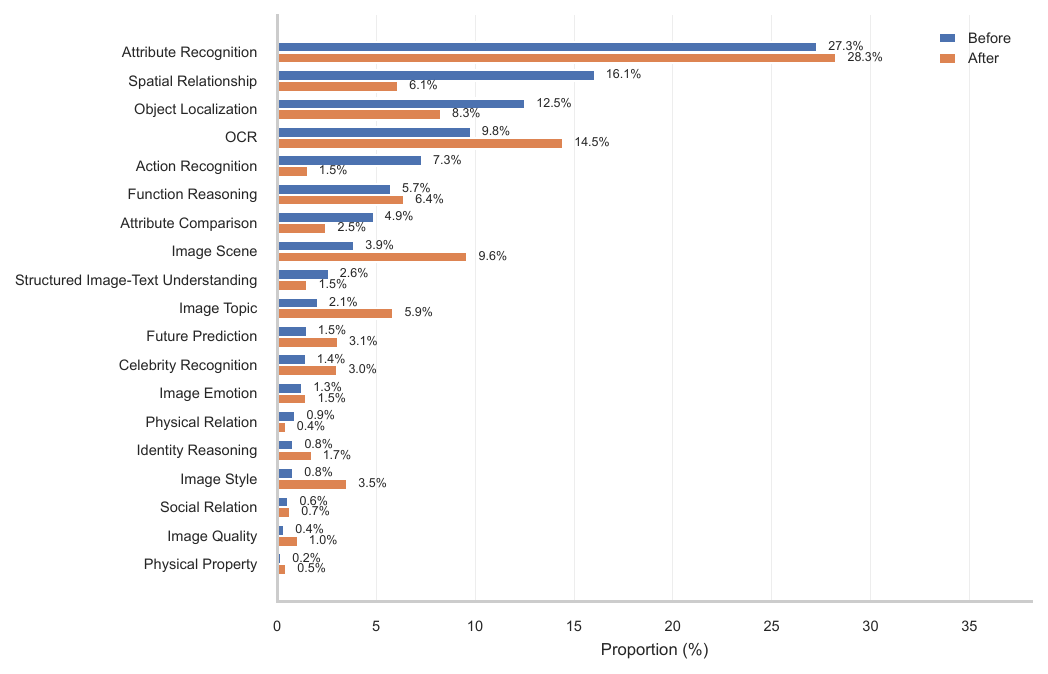}
    \caption{L3-category distribution before and after the redistribution stage.}
    \label{fig:sft_l3_distribution_ws_quad}
\end{figure}

A similar effect is observed at the capability level (Figure~\ref{fig:sft_l3_distribution_ws_quad}). Prior to \emph{Capability Coverage Redistribution}, the distribution shows a clear long-tail, with frequently occurring skills dominating, e.g., Spatial Relationship (16.1\%) and Object Localization (12.5\%). After redistribution, these majority categories are moderated (Spatial Relationship 6.1\% and Object Localization 8.3\%), while sparse but important reasoning abilities are explicitly strengthened, including Future Prediction (1.5\%$\to$3.1\%) and Identity Reasoning(0.8\%$\to$1.7\%). Meanwhile, broad foundational skills are also strengthened (e.g. OCR 9.8\%$\to$14.5\% and Image Topic 2.1\%$\to$5.9\%).

Overall, the \emph{Capability Coverage Redistribution} stage adjusts the sample distribution to mitigate imbalance coverage across both subsets and L3 categories, which is crucial for improving generalization in diverse FSRS scenarios. Notably, a residual long-tail persists, largely due to intrinsic data constraints: categories requiring complex interactions (e.g., Social Relation, Physical Property) remain naturally scarce even after redistributing.

\subsubsection{Tiered Data Organization in OCL.} 
CL provides a systematic framework to organize diverse data formats, including video, text, single-image, and multi-image data. As shown in Table~\ref{tab:ablation_study_on_training_strategy}, jointly optimization on all data types simultaneously leads to a substantial drop in performance. This degradation is likely  due to divergent gradient dynamics between different formats. Moreover, mixing video and images in the same training batch severely degrades training efficiency, as the entire batch must wait for the computationally intensive video processing to complete, creating a major bottleneck.

To resolve these issues, we propose a phased curriculum strategy, as detailed in Table~\ref{tab:data_volume_in_curriculum_learning_training_stage}. Specifically, we introduce a dedicated video-only training stage after Tier 3. By this stage, the model has already learned robust representations of static FSRS images, enabling it to focus exclusively on temporal dynamics without interference from mixed data types within a batch. Following the video stage, Tier 4 incorporates multi-image and text data to broaden the model’s reasoning capabilities across heterogeneous input modalities. This dedicated video-only stage improves both training efficiency and convergence stability.

\begin{table*}[htbp]
\centering
\small
\setlength{\tabcolsep}{6pt}
\renewcommand{\arraystretch}{1.15}
\caption{Data Composition and Volume across OCL Stages.}
\begin{tabular}{lccccc}
\toprule
\textbf{Data Volume} & \textbf{Single Image} & \textbf{MultiImg} & \textbf{Video} & \textbf{Pure Text} \\
\midrule
Tier1 & 1347K & -- & -- & -- \\
Tier2 & 341K & -- & -- & -- \\
Tier3 & 605K & -- & -- & -- \\
Video & -- & -- & 78K & --  \\
Tier4 & 337K & 583K & -- & 25K\\
\bottomrule
\end{tabular}
\label{tab:data_volume_in_curriculum_learning_training_stage}
\end{table*}

\subsubsection{Data Distribution in MPO.}

To further clarify the composition of the MPO stage, we provide a detailed breakdown of the data distribution across distinct scenarios. As shown in Table~\ref{tab:data_in_mpo}, the MPO dataset focuses on three core areas: ShopFront, ShopInterior, and Kitchen. Sample counts are reported in thousands (k) to maintain consistency with the main training tiers.

\begin{table}[htbp]
\caption{MPO Dataset Volume across Scenarios.}
\centering
\small
\setlength{\tabcolsep}{10pt}
\renewcommand{\arraystretch}{1.2}
\begin{tabular}{lcccc}
\toprule
\textbf{Category} & \textbf{ShopFront} & \textbf{ShopInterior} & \textbf{Kitchen} & \textbf{Total} \\
\midrule
Data Volume & 1,609K & 743K  & 862K & 3,214K \\
\bottomrule
\end{tabular}
\label{tab:data_in_mpo}
\end{table}

\subsection{ShopBench Composition}

ShopBench contains a total of 5,818 questions, including 4,716 perception and 1,102 reasoning questions. At Level 2, fine-grained perception dominates, comprising 2,362 single-instance and 1,355 cross-instance questions, while coarse perception accounts for the remaining 999. At Level 3, the most common categories are Object Localization (846), Attribute Recognition (713), and OCR (674), which collectively address core retail capabilities—such as localizing signboards, addresses, and menus, as well as recognizing text in storefront imagery.

ShopBench comprises five internal sub-datasets, reflecting its hybrid organization across scene semantics and visual input formats. Specifically, the single-image portion is divided into three scene-specific subsets: ShopFront, ShopInterior, and Kitchen. In contrast, the multi-image and video data are inherently cross-scenario—each sample may span multiple physical contexts—and are thus treated as two additional, format-defined sub-datasets: MultiImg and Video. The number of questions in each finest-grained category and sub-dataset is reported in the table below.

\begin{figure}[h]
  \centering
  \includegraphics[width=\linewidth]{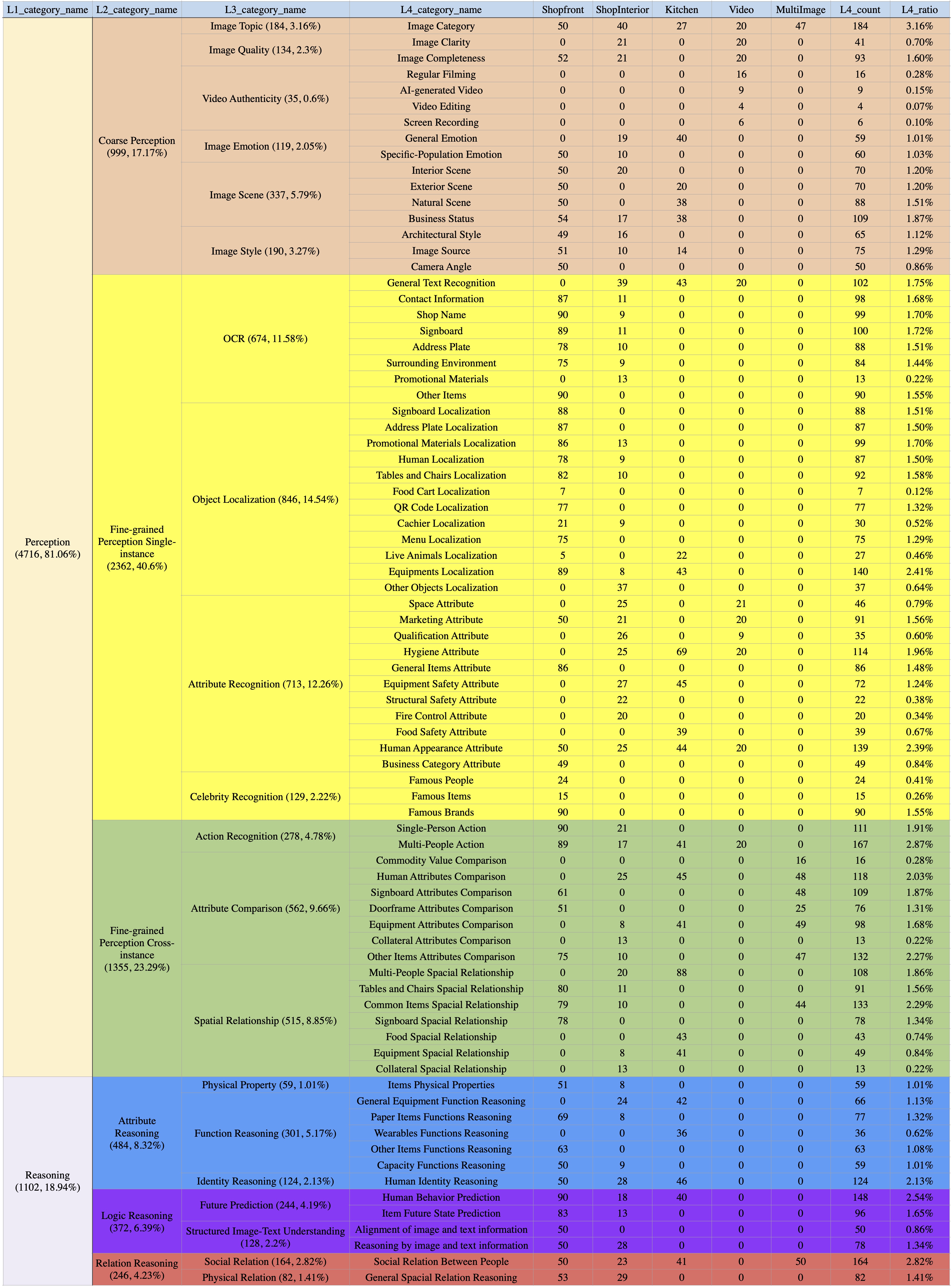}
  \caption{Number of L4 questions in each sub-dataset }
  \label{fig:shopbench_l4_stats}
\end{figure}

\section{An illustrative example of MG and VNR/VIF}
\label{sec:details_of_metrics}

\begin{figure}
    \centering
    \includegraphics[width=1.0\linewidth]{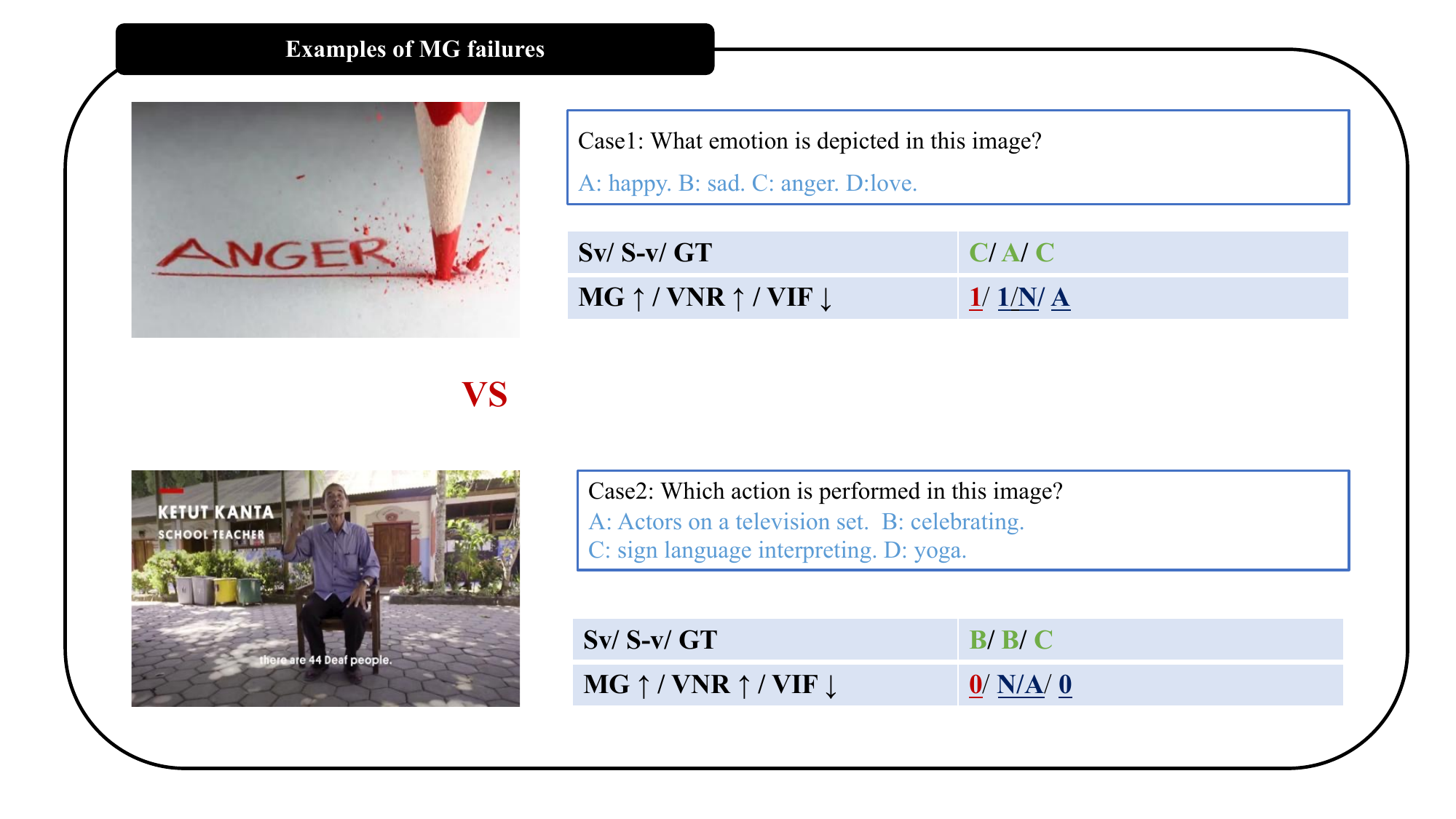}
    \caption{Examples of MG failure, Case2 is a more challenging Question and included in MG computation but MG fails to reflect its soundness.}
    \label{fig:mg_failure}
\end{figure}

In Figure~\ref{fig:mg_failure}, we present an illustrative example: Case 1 represents a straightforward VQA instance, while Case 2 exemplifies a more challenging scenario. MG assigns a score of 1 to Case 1 but 0 to Case 2. This discrepancy highlights MG’s tendency to favor simpler instances, which can distort the perceived difficulty distribution and introduce bias in benchmark construction.

In contrast, both the VNR and VIF offer more balanced and principled numerical characterizations. Specifically, VNR avoids quantifying inherently difficult samples, instead focusing on capturing the presence of meaningful novelty in visual content—thus naturally accommodating simpler instances. Conversely, VIF is designed to account for the cognitive and perceptual complexity in challenging cases by measuring the flow and integration of visual information, while deliberately refraining from assigning significance to overly simplistic samples that require minimal reasoning.

This complementary behavior underscores the advantage of using VNR and VIF in tandem: they provide a more nuanced and equitable assessment across the difficulty spectrum, mitigating the bias observed in MG and enabling a more robust foundation for benchmark curation.

\section{Training and Evaluation Details}
\label{sec:train_eval_details}
\subsection{Training Configuration}
We perform supervised fine-tuning (SFT) of Ostrakon-VL using Qwen3-VL-8B as the foundational model. Training employs full-model fine-tuning for a single epoch, using the AdamW optimizer with a maximum learning rate of $2 \times 10^{-6}$. The learning rate follows a cosine decay schedule with linear warmup over the first 2\% of training steps.
 
To ensure memory efficiency and stable convergence, we use a global batch size of 144. Training employs bfloat16 (BF16) precision and DeepSpeed ZeRO-2 optimization. The maximum sequence length is set to 8192 tokens to accommodate lengthy instruction–response pairs. For visual inputs, images are processed with a maximum resolution of 602,112 pixels (total pixel count). For video inputs, we sample frames at 2 frames per second, capped at 128 frames per video to preserve sufficient temporal context. Detailed hyperparameter settings are provided in Table~\ref{tab:sft_hparams}.

\subsection{Evaluation Protocol}
In our evaluation, we standardize the maximum input resolution across all configurable models to a total of 602,112 pixels for images and 16,777,216 pixels for videos, with proprietary closed-source models as the only exceptions. This alignment mitigates evaluation bias caused by inconsistent input resolutions across benchmarks.

\begin{table}[t]
\centering
\small
\setlength{\tabcolsep}{6pt}
\caption{SFT hyperparameters and training configuration.}
\begin{tabular}{l c}
\toprule
\textbf{Hyperparameter} & \textbf{SFT} \\
\midrule
Base model & Qwen3-VL-8B \\
Fine-tuning type & Full \\
Epochs & 1 \\
Optimizer & AdamW \\
Learning rate & $2\times10^{-6}$ \\
LR schedule & Cosine \\
Warmup ratio & 0.02 \\
Per-device batch size & 1 \\
Gradient accumulation & 2 \\
Global batch size & 144 \\
Max sequence length & 8192 \\
Precision & BF16 \\
DeepSpeed stage & ZeRO-2 \\
FlashAttention & auto \\
Image resolution (pixels) & 602{,}112 \\
Video FPS & 2.0 \\
Max video frames & 128 \\
Video longest edge & 16{,}777{,}216 \\
Video shortest edge & 4{,}096 \\
\bottomrule
\end{tabular}
\label{tab:sft_hparams}
\end{table}

\section{Capability Highlights}
\label{sec:capability_highlights}
This section showcases Ostrakon-VL’s core capabilities that are most critical for real-world merchant risk control and safety inspection workflows. Instead of focusing on a single downstream task, we present results across several fundamental vision-language capabilities, including text understanding (OCR), fine-grained perception (e.g., counting and attribute recognition), multi-image consistency reasoning, and video understanding (e.g., extracting temporal evidence and detecting AI-generated videos). All examples are drawn from real inspection scenarios and evaluated under standardized prompts and consistent decision criteria, enabling efficient human review and actionable escalation in deployment.

\begin{figure}[htbp]
  \centering
\includegraphics[width=\linewidth,trim=1cm 1.7cm 1cm 0.7cm,clip]{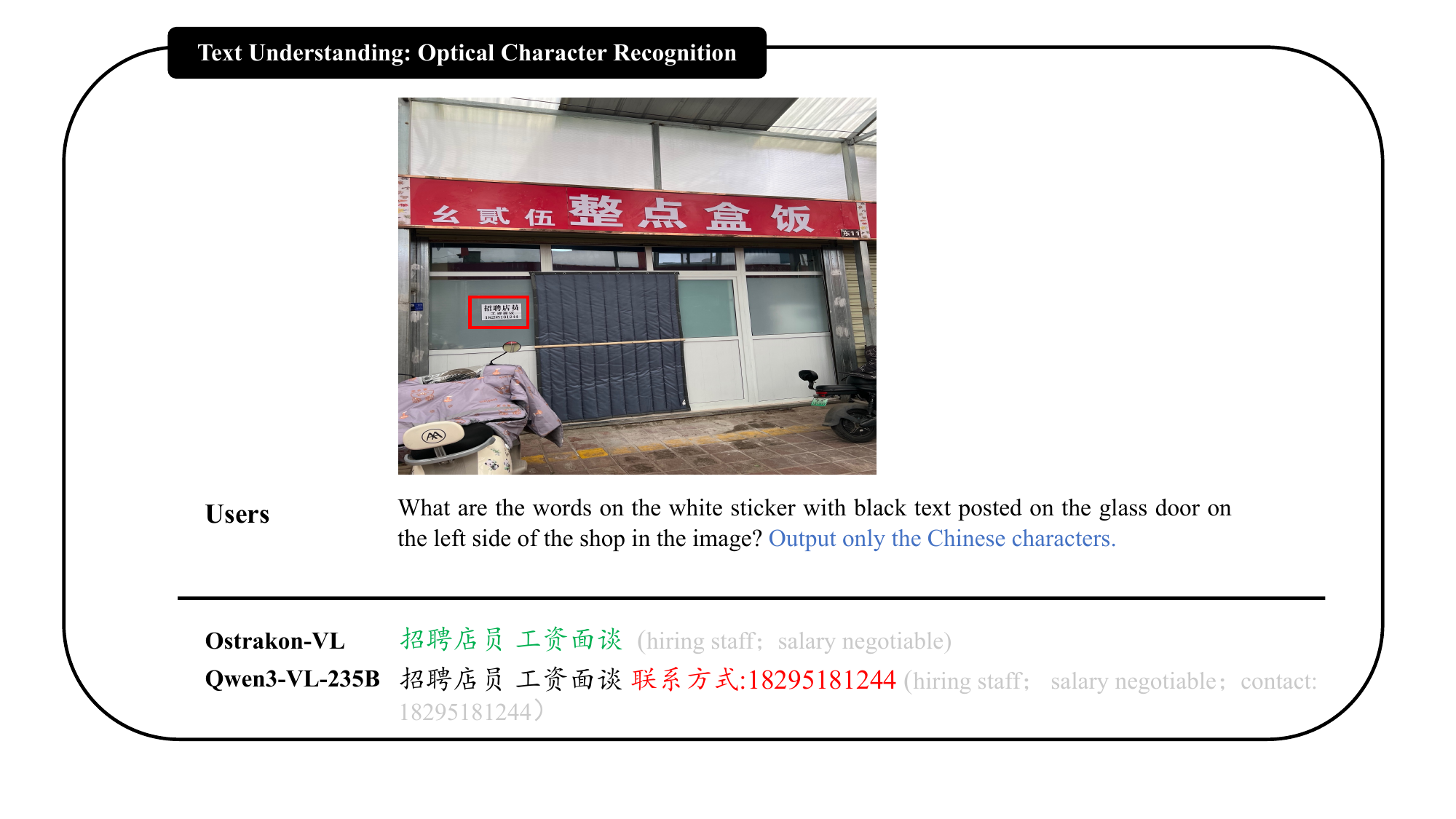}
  \caption{Text  Understanding (Optical Character Recognition). The model is prompted to read the black text printed on a white sticker posted on a storefront glass door, extracting the recruitment notice content under real inspection conditions.}
  \label{fig:ocr_case}
\end{figure}

\subsection{Text Understanding}
\begin{figure}[htbp]
  \centering
\includegraphics[width=\linewidth,trim=1cm 1.7cm 1cm 0.7cm,clip]{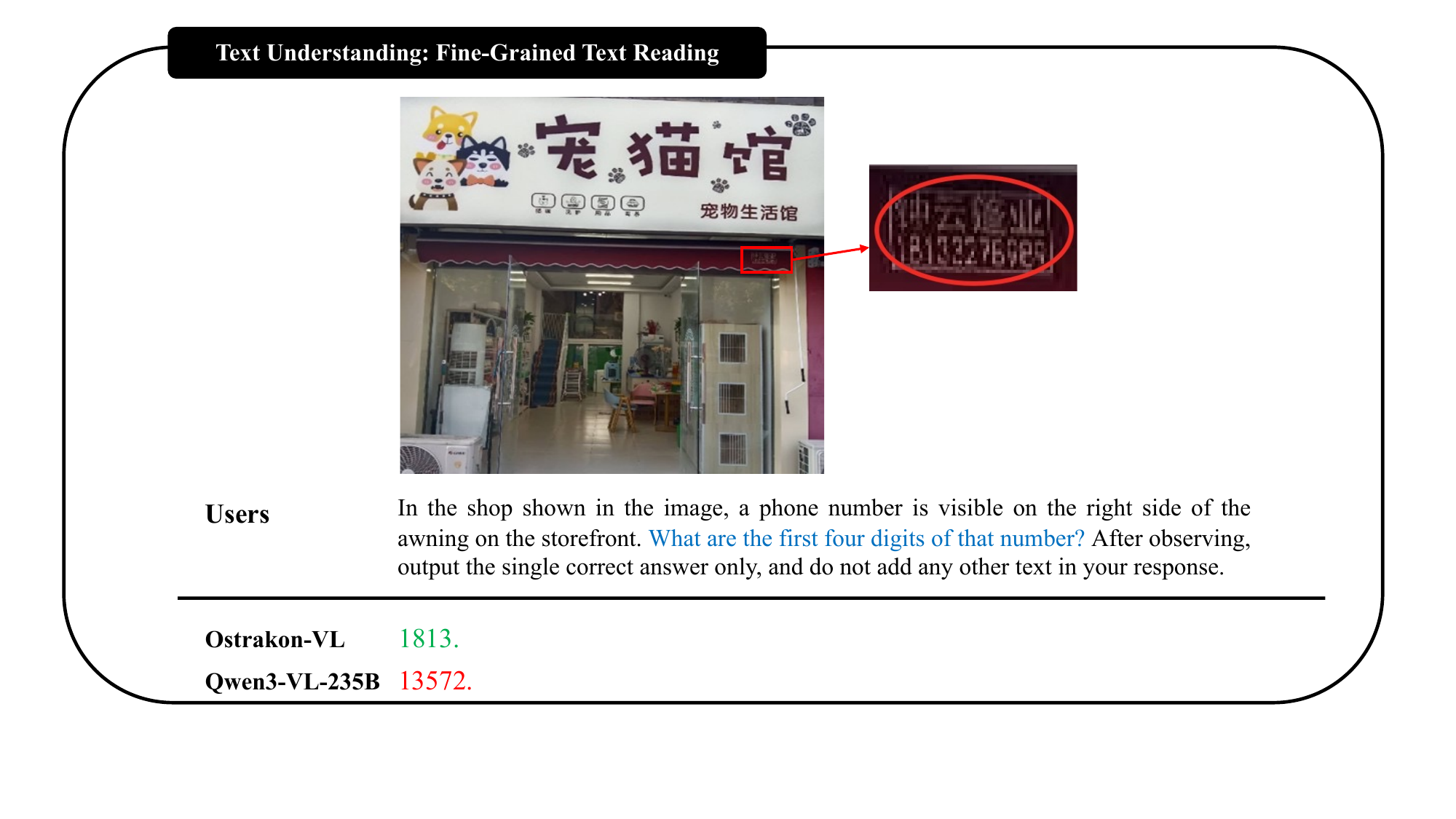}
  \caption{Text Understanding (Fine-Grained Text Reading). The model must locate a phone number printed on the storefront awning area and output only the first four digits, testing digit-level precision and constraint adherence.}
  \label{fig:fine_text_reading_case}
\end{figure}

Text understanding is a cornerstone of merchant inspection: models must reliably recognize storefront signage, posted notices, and fine-print contact information under non-ideal capture conditions such as glare, perspective distortion, partial occlusion, and cluttered backgrounds. We evaluate Ostrakon-VL on two representative capabilities commonly encountered in real-world workflows: (i) robust OCR for extracting semantic content from in-the-wild notices, and (ii) fine-grained text recognition requiring precise character-level extraction—including digits—under strict output-only constraints. As shown in Figures~\ref{fig:ocr_case} and~\ref{fig:fine_text_reading_case}, Ostrakon-VL adheres to instruction-specified output formats (e.g., output-only requirements) while maintaining high fidelity to the visible text.

\subsection{Fine-grained Visual Understanding}
\begin{figure}[htbp]
  \centering
    \centering
\includegraphics[width=\linewidth,trim=1cm 2.5cm 1cm 0.7cm,clip]{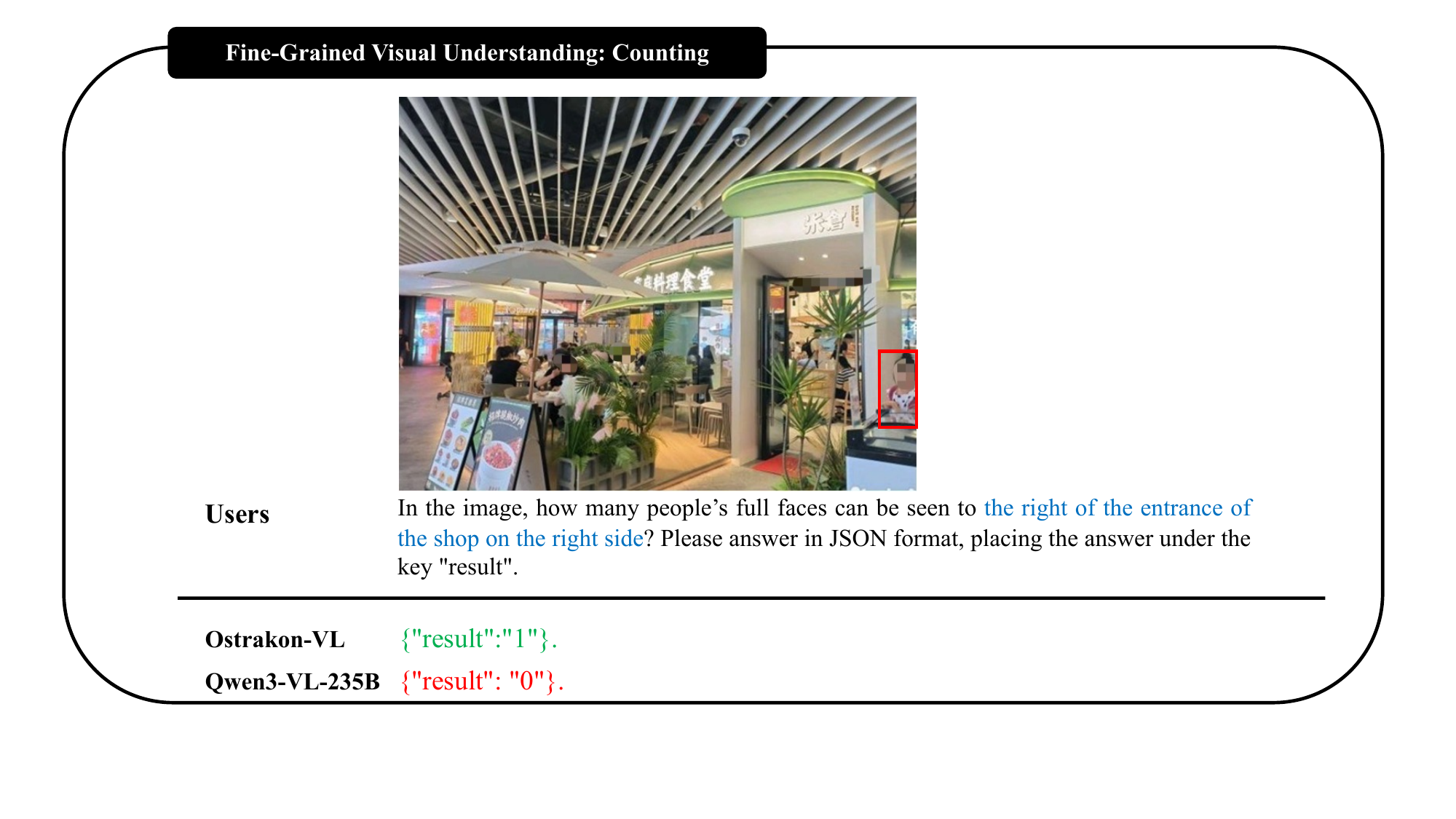}
  \caption{Fine-Grained Visual Understanding (Counting). A complementary counting instance evaluated with the same protocol, emphasizing region-conditioned counting and strict adherence to structured response formatting.}
  \label{fig:counting_case_2}
\end{figure}

Fine-grained perception enables critical inspection capabilities such as localized counting, region-specific evidence retrieval, and attribute recognition. We evaluate Ostrakon-VL on spatially grounded counting tasks that require reasoning under explicit positional constraints (e.g., a specified side of an entrance) and producing structured outputs. Figure~\ref{fig:counting_case_2} illustrates that the model aligns its predictions with the queried region and returns results in a JSON format conforming to the required schema, demonstrating reliable spatial grounding rather than reliance on global scene heuristics.

\begin{figure}[htbp]
  \centering
    \centering
\includegraphics[width=\linewidth,trim=1cm 2.3cm 1cm 0.7cm,clip]{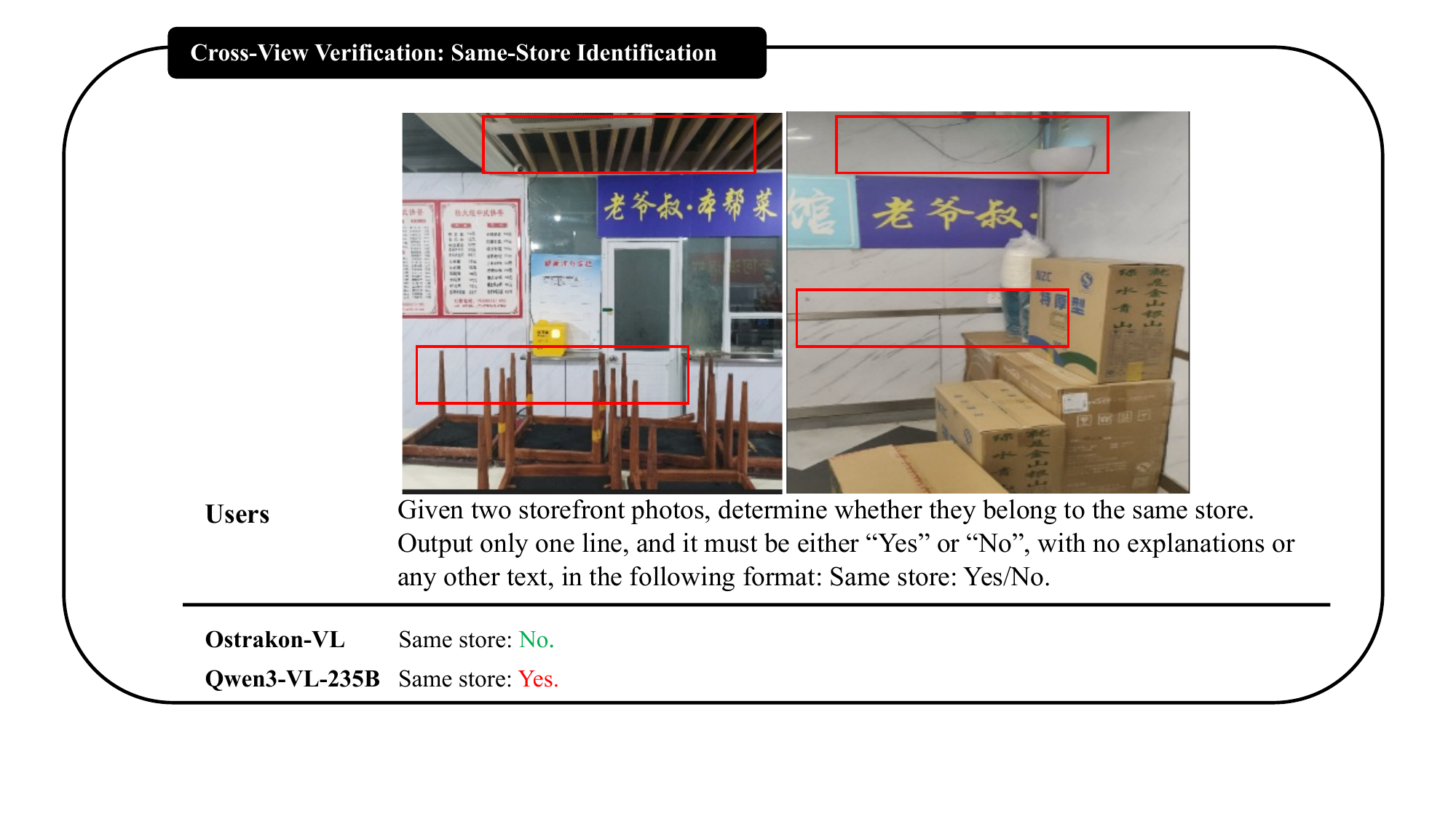}
  \caption{Cross-View Verification (Same-Store Identification). Although the two storefront images share some similar structural cues, key immutable architectural features (e.g., the ceiling structure, wall divider lines, and spatial layout) cannot be consistently aligned across views; therefore, they are judged to be different stores.}
  \label{fig:same_store_identification_case}
\end{figure}

\subsection{Multi-image Perception}

\begin{figure}[t]
  \centering
    \centering
\includegraphics[width=\linewidth,trim=1cm 2.3cm 0.7cm 0.7cm,clip]{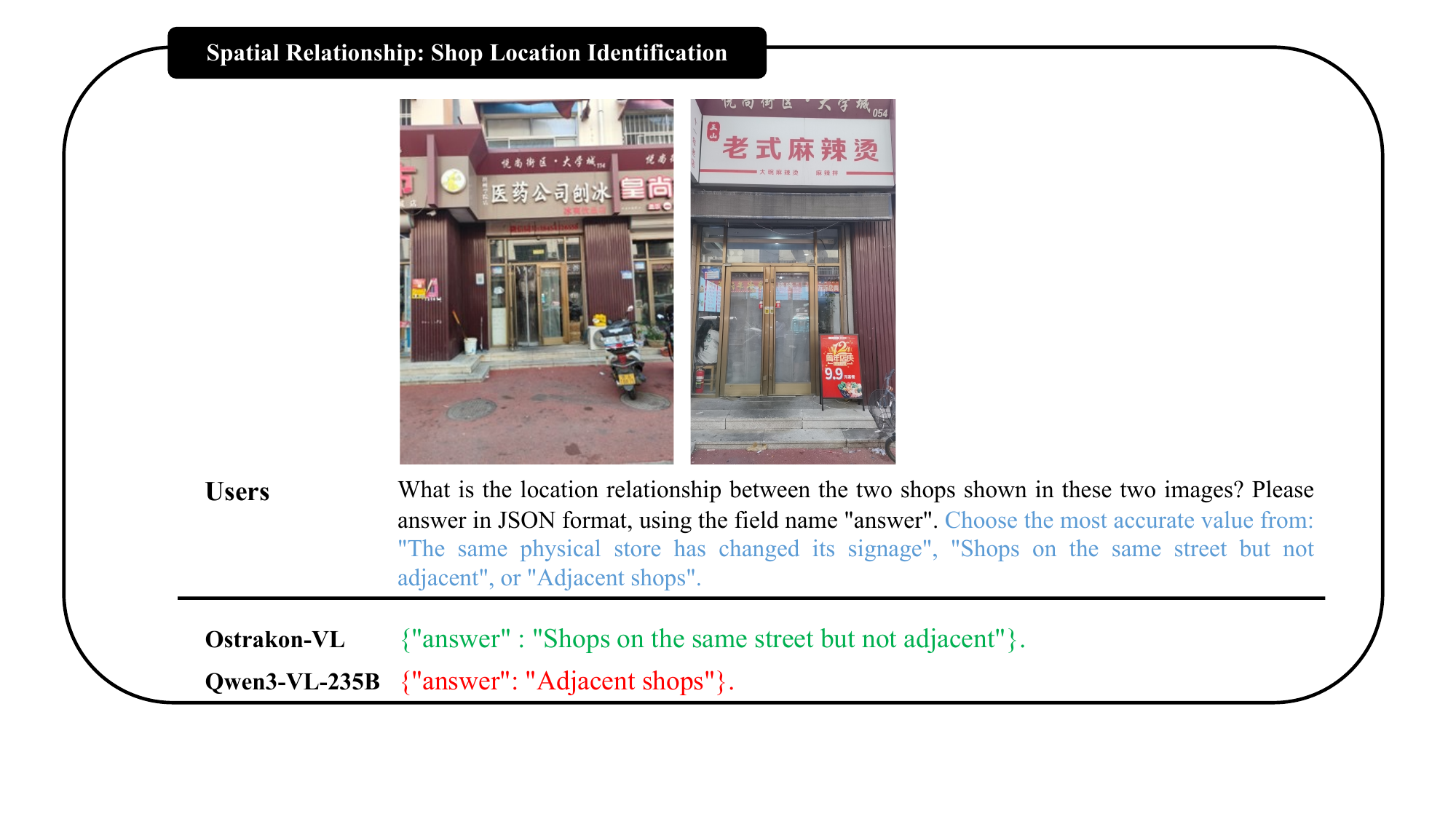}
  \caption{Spatial Relationship (Shop Location Identification). Given two storefront photos, the model predicts the location relationship using a fixed label set (same physical store with changed signage / same street but not adjacent / adjacent), grounding the decision in cross-image architectural and contextual cues.}
  \label{fig:shop_location_relationship_case}
\end{figure}

Inspection workflows commonly require cross-image perception to determine whether two photos depict the same physical storefront (possibly captured at different times) or to assess the spatial relationship between two shops (e.g., adjacent vs.\ non-adjacent on the same street). We evaluate Ostrakon-VL using paired-image prompts and a constrained decision space, requiring the model to rely on stable geometric and contextual cues (e.g., facade layout, door configuration, surrounding architectural elements) rather than transient visual attributes such as textual content on signage. Figure~\ref{fig:same_store_identification_case} demonstrates same-store verification under significant appearance changes, while Figure~\ref{fig:shop_location_relationship_case} shows the model’s ability to infer categorical spatial relationships from pairs of storefront images.

\subsection{Video Understanding}
\begin{figure}[htbp]
  \centering
\includegraphics[width=\linewidth,trim=1cm 0.7cm 1cm 0.7cm,clip]{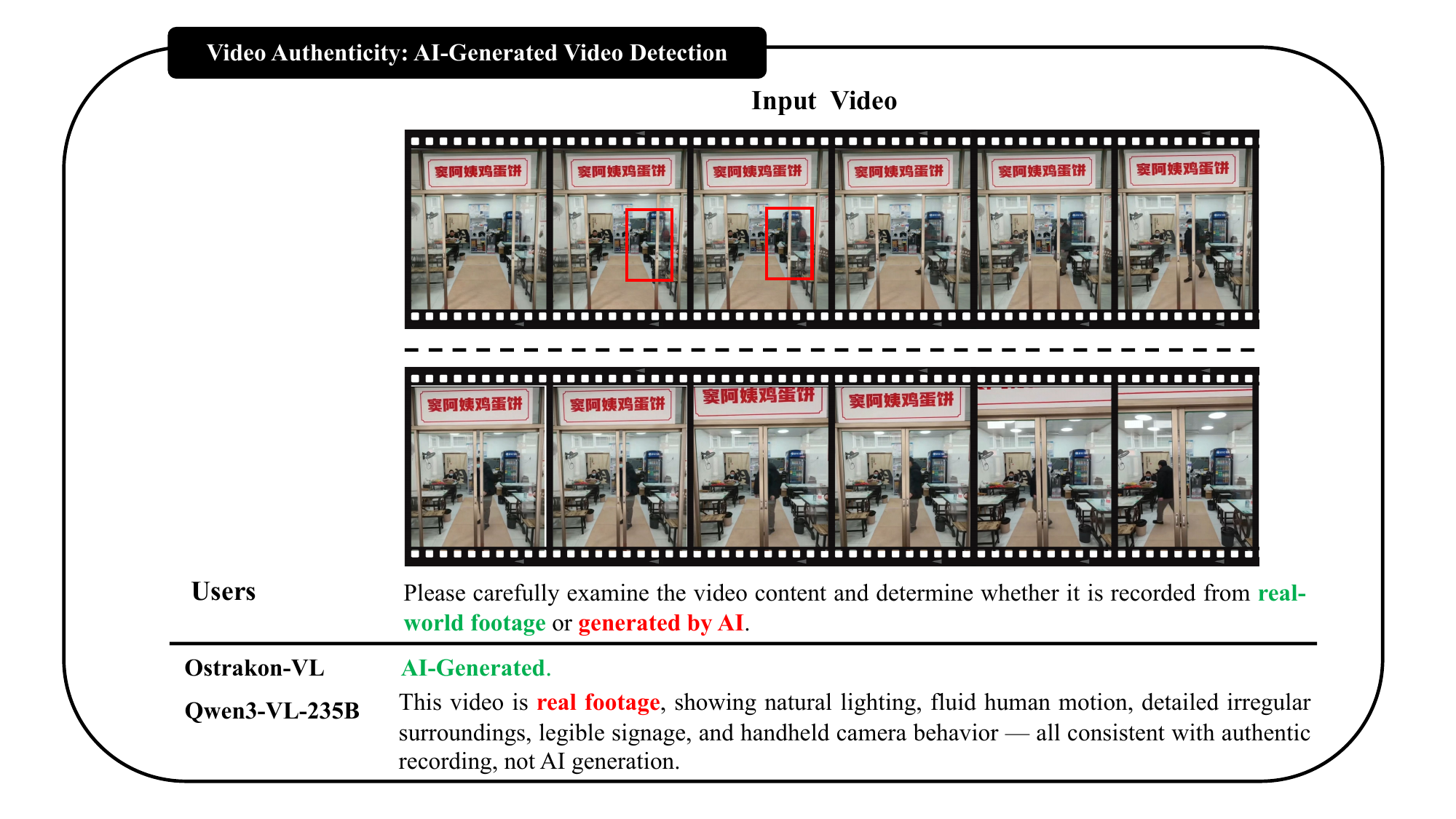}
  \caption{Video Authenticity (AI-Generated Video Detection). Uniformly sampled frames show discontinuities in the person’s position across adjacent frames and abnormal local edge and texture details; therefore, the video is judged to be AI-Generated.}
  \label{fig:aigv_detection_case}
\end{figure}

We evaluate the detection of AI-generated videos in risk control scenarios where fabricated evidence may appear during remote inspections. For each clip, we uniformly sample representative frames and pose a standardized binary query asking whether the video is real-world footage or AI-generated. Figure~\ref{fig:aigv_detection_case} shows an example where the input comprises multiple frames from an indoor retail scene. The model must base its judgment on temporal and cross-frame consistency, scene realism, and the plausibility of fine visual details to produce a binary authenticity decision.

\subsection{Risk and Compliance Reasoning}
\begin{figure}[t]
   \centering
\includegraphics[width=\linewidth,trim=1cm 2.1cm 1cm 0.7cm,clip]{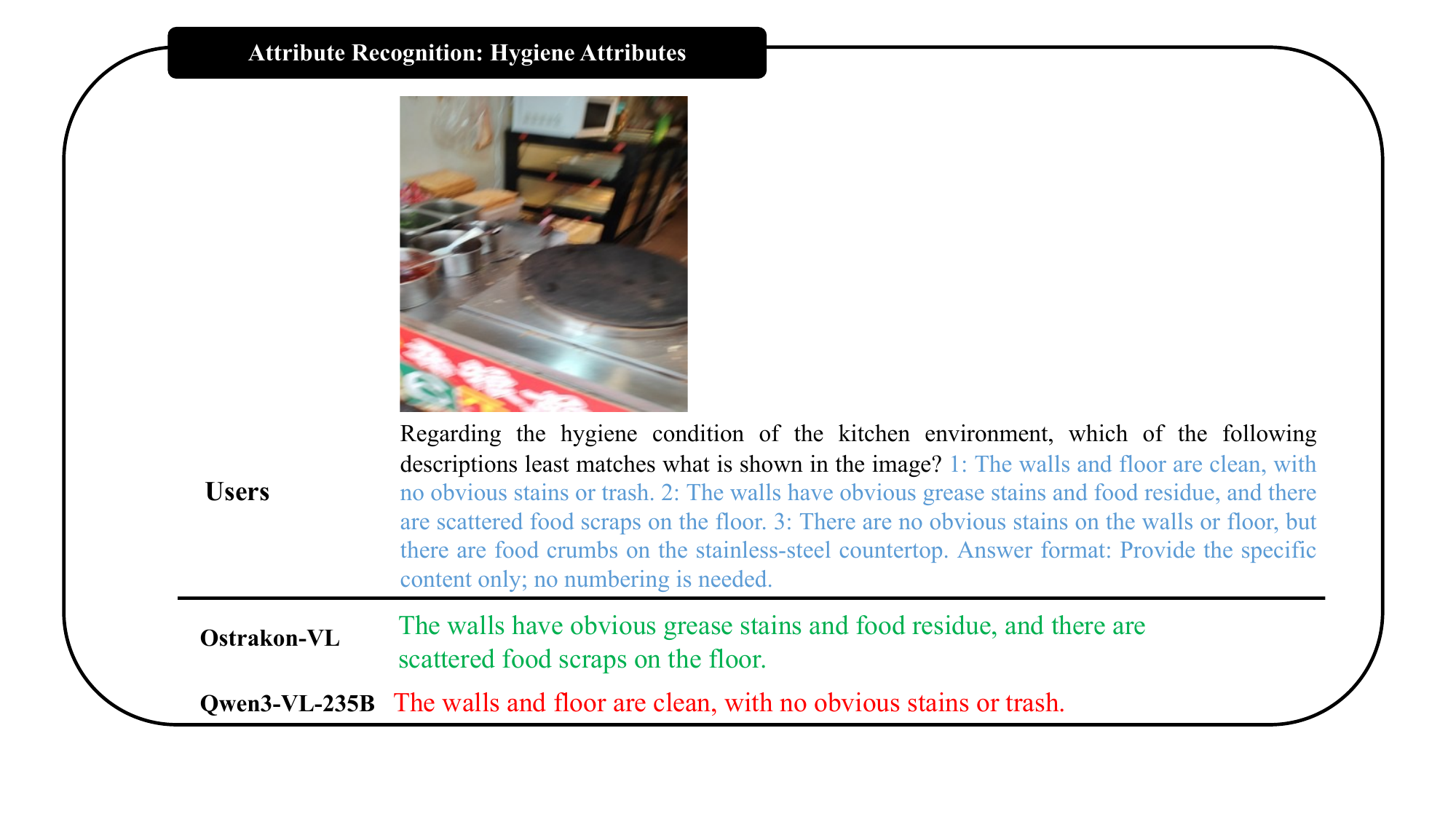}
  \caption{Attribute Recognition (Hygiene Attributes). The model is asked to select, from multiple textual statements, the one that least matches the observed kitchen scene, requiring evidence-based judgment consistent with standardized inspection criteria.}
  \label{fig:hygiene_attribute_case}
\end{figure}

In real-world risk control and compliance auditing, models must derive actionable, policy-compliant conclusions from perceptual evidence. This requires (i) grounded decision-making (linking conclusions to observable cues), (ii) calibrated behavior under uncertainty (avoiding overconfident predictions when evidence is insufficient), and (iii) consistent and robust performance across heterogeneous inputs under predefined decision protocols. We evaluate Ostrakon-VL using unified prompts and fixed criteria that mirror real-world deployment practices, across tasks spanning textual evidence extraction, location verification, and safety-related assessment. Figure~\ref{fig:hygiene_attribute_case} illustrates compliance-oriented reasoning in a kitchen setting: the model selects the candidate description that is least consistent with the visual evidence—a task formulation commonly adopted in audit checklists and escalation pipelines.

\end{document}